\documentclass[10pt,journal,compsoc]{IEEEtran}
\usepackage{cite}

\ifCLASSINFOpdf
\usepackage[pdftex]{graphicx}
\else
\fi
\usepackage{amsmath}
\usepackage{algorithmic}
\usepackage{multirow}
\usepackage{threeparttable}
\usepackage{amssymb}
\usepackage{makecell}
\usepackage{booktabs}

\usepackage{algorithm} 
\usepackage{algorithmic}
\usepackage{threeparttable}
\usepackage{bbm}
\usepackage{hyperref}
\usepackage{subfig}  
\usepackage{url}
\usepackage{balance}
\usepackage{enumitem}
\usepackage{soul, color, xcolor}
\begin{document}
%

\title{Self-Supervised Learning for Time Series Analysis: Taxonomy, Progress, and Prospects}

%
%
%
\author{
        Kexin Zhang,
        Qingsong Wen$^\dag$,
        Chaoli Zhang,
        Rongyao Cai,
        Ming Jin,
        Yong Liu$^\dag$,
        James Y. Zhang,\\
        Yuxuan Liang,
        Guansong Pang,
        Dongjin Song,
        and Shirui Pan

\thanks{Kexin Zhang, Rongyao Cai, and Yong Liu are with the Institute of Cyber-Systems and Control, Zhejiang University. (e-mail: zhangkexin@zju.edu.cn, rycai@zju.edu.cn, and yongliu@iipc.zju.edu.cn)}
\thanks{Qingsong Wen is with Squirrel AI, and this work was done at Alibaba Group. (e-mail: qingsongedu@gmail.com)}
\thanks{Chaoli Zhang is with the School of Computer Science and Technology, Zhejiang Normal University. (e-mail: chaolizcl@zjnu.edu.cn)}
\thanks{Ming Jin is with the Faculty of Information Technology, Monash University. (e-mail: ming.jin@monash.edu)}
\thanks{James Zhang is with Ant Group. (e-mail: james.z@antgroup.com)}
\thanks{Yuxuan Liang is with INTR \& DSA Thrust, Hong Kong University of Science and Technology (Guangzhou). (e-mail: yuxliang@outlook.com)}
\thanks{Guansong Pang is with the School of Computing and Information Systems, Singapore Management University. (e-mail: gspang@smu.edu.sg)}
\thanks{Dongjin Song is with the School of Computing, University of Connecticut. (e-mail: dongjin.song@uconn.edu)}
\thanks{Shirui Pan is with the School of Information and Communication Technology, Griffith University. (e-mail: s.pan@griffith.edu.au)}
\thanks{$^\dag$Corresponding author: Qingsong Wen and Yong Liu.}
\thanks{GitHub Page: \url{https://github.com/qingsongedu/Awesome-SSL4TS}}
}


%
%

\markboth{IEEE Transactions on Pattern Analysis and Machine Intelligence}{}

%



\IEEEtitleabstractindextext{%
\begin{abstract}
Self-supervised learning (SSL) has recently achieved impressive performance on various time series tasks. The most prominent advantage of SSL is that it reduces the dependence on labeled data. Based on the pre-training and fine-tuning strategy, even a small amount of labeled data can achieve high performance. Compared with many published self-supervised surveys on computer vision and natural language processing, a comprehensive survey for time series SSL is still missing. To fill this gap, we review current state-of-the-art SSL methods for time series data in this article. To this end, we first comprehensively review existing surveys related to SSL and time series, and then provide a new taxonomy of existing time series SSL methods by summarizing them from three perspectives: generative-based, contrastive-based, and adversarial-based. These methods are further divided into ten subcategories with detailed reviews and discussions about their key intuitions, main frameworks, advantages and disadvantages. To facilitate the experiments and validation of time series SSL methods, we also summarize datasets commonly used in time series forecasting, classification, anomaly detection, and clustering tasks. Finally, we present the future directions of SSL for time series analysis. 
\end{abstract}

\begin{IEEEkeywords}
Time series analysis, self-supervised learning, representation learning, deep learning
\end{IEEEkeywords}}

\maketitle

\IEEEdisplaynontitleabstractindextext

%
\IEEEpeerreviewmaketitle

\newcommand{\tabincell}[2]{\begin{tabular}{@{}#1@{}}#2\end{tabular}}





\section{Introduction}
%
%
%
%


\IEEEPARstart{T}{ime} series data abound in many real-world scenarios~\cite{wen2022robust,esling2012time}, including human activity recognition \cite{10.5555/2832747.2832806}, industrial fault diagnosis \cite{9448494}, smart building management \cite{Li_Hong_Wang_2020}, and healthcare \cite{10.1145/3219819.3220051}. The key to most tasks based on time series analysis is to extract useful and informative features. In recent years, Deep Learning (DL) has shown impressive performance in extracting hidden patterns and features of the data. Generally, the availability of sufficiently large labeled data is one of the critical factors for a reliable DL-based feature extraction model, usually referred to as supervised learning. Unfortunately, this requirement is difficult to meet in some practical scenarios, particularly for time series data, where obtaining labeled data is a time-consuming process. As an alternative, Self-Supervised Learning (SSL) has garnered  
increasing 
attention for its label-efficiency and generalization ability, and consequently, many latest time series modeling methods have been following this learning paradigm.

SSL is a subset of unsupervised learning that utilizes pretext tasks to derive supervision signals from unlabeled data. These pretext tasks are self-generated challenges that the model solves to learn from the data, thereby creating valuable representations for downstream tasks.
SSL does not require additional manually labeled data because the supervisory signal is derived from the data itself. With the help of well-designed pretext tasks, SSL has recently achieved great success in the domains of Computer Vision (CV) \cite{gidaris2018unsupervised,8099559,NIPS2014_07563a3f,10.1007/978-3-030-58621-8_45} and Natural Language Processing (NLP) \cite{devlin-etal-2019-bert,gao2021simcse}. 

\begin{figure*}[!t]
	\centering
	\includegraphics[width=175mm]{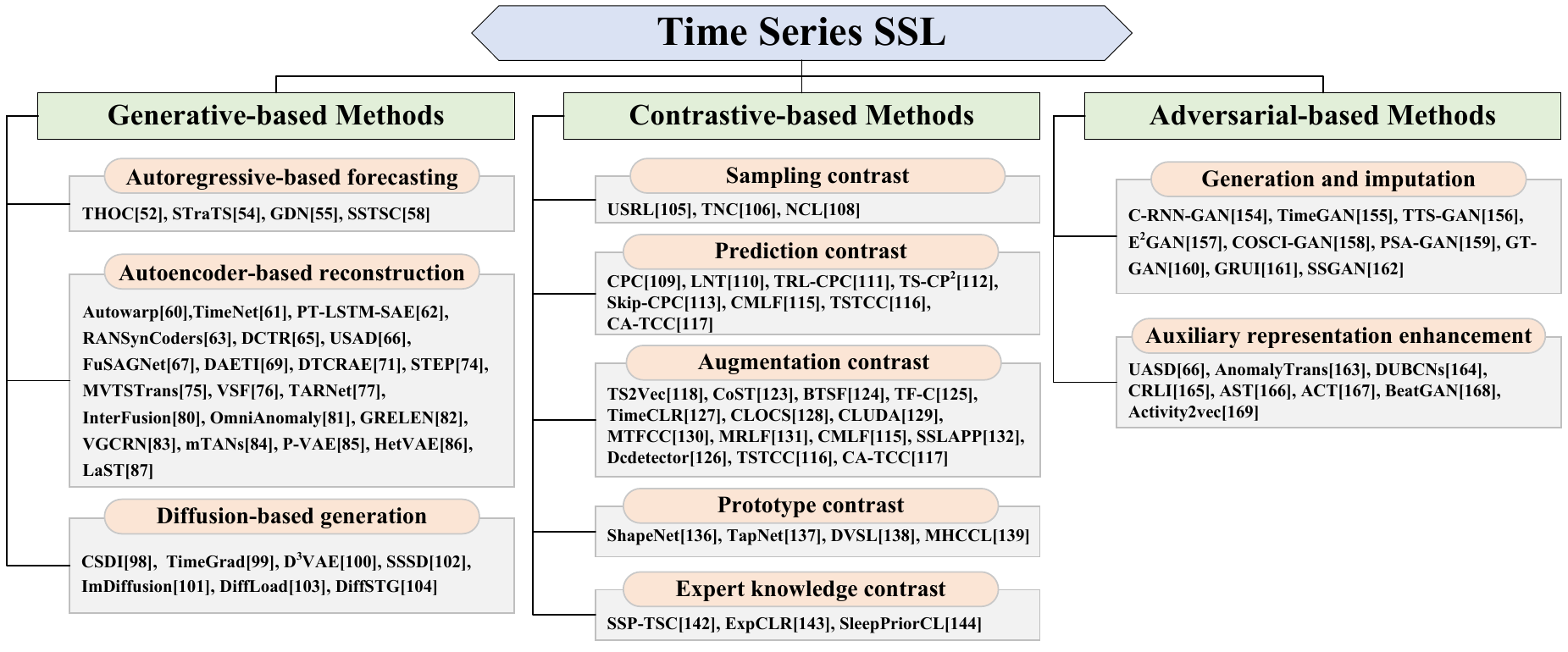}
	\caption{The proposed taxonomy of SSL for time series data.}
	\label{fig:taxonomy}
    \vspace{-10pt}
\end{figure*}

With the great success of SSL in CV and NLP, it is appealing 
to extend SSL to time series data. However, transferring the pretext tasks designed for CV/NLP directly to time series data
 is non-trivial, and often fails to work in many scenarios.
 Here we highlight some typical challenges that arise when applying SSL to time series data. First, time series data exhibit unique properties such as seasonality, trend, and frequency domain information~\cite{cleveland1990stl,wen2020fast,zhou2022film}. Since most pretext tasks designed for image or language data do not consider these semantics related to time series data, they cannot be directly adopted. Second, some techniques commonly used in SSL, such as data augmentation, need to be specially designed for time series data. For example, rotation and crop are the commonly used augmentation techniques for image data \cite{10.5555/3524938.3525087}. However, these two techniques may break the temporal dependency of the series data. Third,  most time series data contain multiple dimensions, i.e., multivariate time series. However, useful information usually only exists in a few dimensions, making it difficult to extract useful information in time series using SSL methods from other data types. 


To the best of our knowledge, there has yet to be a comprehensive and systematic review of SSL for time series data, in contrast to the extensive literature on SSL for CV or NLP~\cite{technologies9010002,9086055}. The surveys proposed by Eldele et al. \cite{eldele2023labelefficient} and Deldari et al. \cite{DBLP:journals/corr/abs-2206-02353} are partly similar to our work. However, these two reviews only discuss a small part of self-supervised contrastive learning (SSCL), which requires a more comprehensive literature review. Furthermore, a summary of benchmark time series datasets needs to be included, and the potential research directions for time series SSL are also scarce.

This article provides a review of current state-of-the-art SSL methods for time series data. We begin by summarizing recent reviews on SSL and time series data and then propose a new taxonomy from three perspectives: generative-based, contrastive-based, and adversarial-based. The taxonomy is similar to the one proposed by Liu et al. \cite{9462394} but specifically concentrated on time series data. For generative-based methods, we describe three frameworks: autoregressive-based forecasting, auto-encoder-based reconstruction, and diffusion-based generation. For contrastive-based methods, we divide the existing work into five categories based on 
how positive and negative samples are generated, including sampling contrast, prediction contrast, augmentation contrast, prototype contrast, and expert knowledge contrast. Then we sort out and summarize the adversarial-based methods based on two target tasks: time series generation/imputation and auxiliary representation enhancement. The proposed taxonomy is shown in Fig. \ref{fig:taxonomy}. We conclude this work by discussing possible future directions for time series SSL, including selection and combination of data augmentation, selection of positive and negative samples in SSCL, the inductive bias for time series SSL, theoretical analysis of SSCL, adversarial attacks and robust analysis on time series, time series domain adaption, pretraining and large models for time series, time series SSL in collaborative systems, and benchmark evaluation for time series SSL. 

Our main contributions are summarized as follows. 
\begin{itemize}[leftmargin=*]
    \item \textbf{New taxonomy and comprehensive review.} We provide a new taxonomy and a detailed and up-to-date review of time series SSL. We divide existing methods into ten categories, and for each category, we describe the basic frameworks, mathematical expression, fine-grained classification, detailed comparison, advantages and disadvantages. To the best of our knowledge, this is the first work to comprehensively and systematically review the existing studies of SSL for time series data.
    
    \item \textbf{Collection of applications and datasets.} We collect resources on time series SSL, including applications and datasets, and investigate related data sources, characteristics, and corresponding works.
    
    \item \textbf{Abundant future directions.} We point out key problems in this field from both applicative and methodology perspectives, analyze their causes and possible solutions, and discuss future research directions for time series SSL. We strongly believe that our efforts will ignite further research interests in time series SSL.

\end{itemize}

The rest of the article is organized as follows. Section \ref{sec:relatedsurveys} provides some review literature on SSL and time series data. Section \ref{sec:generative} to Section \ref{sec:adversarial}  describe the generation-based, contrastive-based, and adversarial-based methods, respectively. Section \ref{sec:applications} lists some commonly used time series data sets from the application perspective. The quantitative performance comparisons and discussions are also provided. Section \ref{sec:future} discusses promising directions of time series SSL, and Section \ref{sec:con} concludes the article.

\section{Related Surveys}
\label{sec:relatedsurveys}
In this section, the definition of time series data is first introduced, and then several recent reviews on SSL and time series analysis are scrutinized.

\subsection{Definition of time series data}

\subsubsection{Univariate time series}
A univariate time series refers to an ordered sequence of observations or measurements of the same variable indexed by time. It can be defined as $X = (x_0, x_1, x_2, \ldots, x_t)$,  where $x_i$ is the point at timestamp $i$. Most often, the measurements are made at regular time intervals.

\subsubsection{Multivariate time series}
A multivariate time series consists of two or more interrelated variables (or dimensions) that depend on time. It is a combination of multiple univariate time series and can be defined as $\mathbf{X} = \left[ X_0, X_1, X_2, \ldots, X_p \right]$, where $p$ is the number of variables.

\subsubsection{Multiple multivariate time series}
Considering the scenario where distinct sets of multivariate time series are concurrently examined. Analyzing such datasets involves studying each set independently and exploring the relationships between different sets. For instance, if we study meteorological data from different cities, each city's data forms a multivariate time series, collectively resulting in multiple multivariate time series. This can be articulated as $\mathcal{X} = \left\{\mathbf{X}_0, \mathbf{X}_1, \mathbf{X}_2, \ldots, \mathbf{X}_n \right\}$, where $n$ is the number of multivariate time series.

\subsection{Surveys on SSL}

The surveys on SSL can be categorized by different criteria. In this paper, we outline three widely used criteria: learning paradigms, pretext tasks and components/modules. 

\subsubsection{Learning paradigms} This category focuses on model architectures and training objectives. The SSL methods can be roughly divided into the following categories: generative-based, contrastive-based, and adversarial-based methods. The characteristics and descriptions of the above methods can be found in Appendix~\ref{app:lp}. Using the learning paradigm as a taxonomy is arguably the most popular among the existing SSL surveys, including \cite{6472238,DBLP:journals/corr/abs-2206-02353,https://doi.org/10.48550/arxiv.2203.01205,9893562,9770382,9764632,DBLP:journals/corr/abs-2003-08271}. However, not all surveys cover the above three categories. The readers are referred to these surveys for more details. In Table \ref{table-sslsurveys}, we also provide the data modalities involved in each survey, which can help readers quickly find the research work closely related to them.


    
    
    
    


\begin{table}[!t]
	\caption{An overview of recent SSL surveys on different modalities. The yellow part indicates the proportion of this modality in all six modalities. It can be seen that this article is mainly concerned with SSL on time series data.}
	\label{table-sslsurveys}
	\centering
	\begin{tabular}{c|cccccc} \toprule\toprule
	
	\multirow{2}{*}{Paper} & \multicolumn{6}{c}{Modality} \\  \cmidrule(lr){2-7}
	
	&Image &Video & Audio & Graph & Text & Time Series \\ \midrule
 
    \cite{9462394} &
    $\includegraphics[width=5mm]{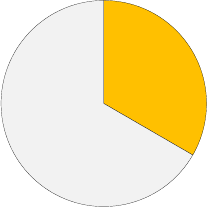}$ & $\includegraphics[width=5mm]{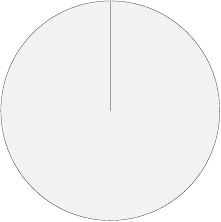}$&
    $\includegraphics[width=5mm]{figures/mod_13.pdf}$&
    $\includegraphics[width=5mm]{figures/mod_13.pdf}$ & $\includegraphics[width=5mm]{figures/mod_0.pdf}$ & $\includegraphics[width=5mm]{figures/mod_0.pdf}$ \\ \midrule 
    
    \cite{6472238} &
    $\includegraphics[width=5mm]{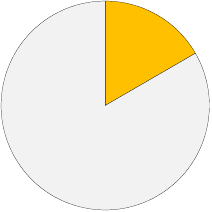}$ &
    $\includegraphics[width=5mm]{figures/mod_16.pdf}$&
    $\includegraphics[width=5mm]{figures/mod_16.pdf}$&
    $\includegraphics[width=5mm]{figures/mod_16.pdf}$&
    $\includegraphics[width=5mm]{figures/mod_16.pdf}$&
    $\includegraphics[width=5mm]{figures/mod_16.pdf}$ \\ \midrule 
    
    \cite{9770283} & 
    $\includegraphics[width=5mm]{figures/mod_16.pdf}$ &
    $\includegraphics[width=5mm]{figures/mod_16.pdf}$&
    $\includegraphics[width=5mm]{figures/mod_16.pdf}$&
    $\includegraphics[width=5mm]{figures/mod_16.pdf}$&
    $\includegraphics[width=5mm]{figures/mod_16.pdf}$&
    $\includegraphics[width=5mm]{figures/mod_16.pdf}$ \\ \midrule 
    
    \cite{9226466} & 
    $\includegraphics[width=5mm]{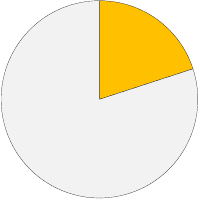}$ &
    $\includegraphics[width=5mm]{figures/mod_15.pdf}$&
    $\includegraphics[width=5mm]{figures/mod_15.pdf}$&
    $\includegraphics[width=5mm]{figures/mod_15.pdf}$&
    $\includegraphics[width=5mm]{figures/mod_15.pdf}$& 
    $\includegraphics[width=5mm]{figures/mod_0.pdf}$ \\ \midrule 
    
    \cite{technologies9010002} & $\includegraphics[width=5mm]{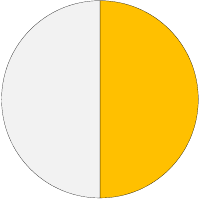}$ &$\includegraphics[width=5mm]{figures/mod_0.pdf}$&$\includegraphics[width=5mm]{figures/mod_0.pdf}$&$\includegraphics[width=5mm]{figures/mod_0.pdf}$&$\includegraphics[width=5mm]{figures/mod_12.pdf}$&$\includegraphics[width=5mm]{figures/mod_0.pdf}$ \\ \midrule 
    
    \cite{DBLP:journals/corr/abs-2206-02353}&  $\includegraphics[width=5mm]{figures/mod_15.pdf}$ &$\includegraphics[width=5mm]{figures/mod_15.pdf}$&$\includegraphics[width=5mm]{figures/mod_15.pdf}$&$\includegraphics[width=5mm]{figures/mod_0.pdf}$&$\includegraphics[width=5mm]{figures/mod_15.pdf}$& $\includegraphics[width=5mm]{figures/mod_15.pdf}$ \\ \midrule 
    
    \cite{9086055} & $\includegraphics[width=5mm]{figures/mod_12.pdf}$&$\includegraphics[width=5mm]{figures/mod_12.pdf}$&$\includegraphics[width=5mm]{figures/mod_0.pdf}$&$\includegraphics[width=5mm]{figures/mod_0.pdf}$&$\includegraphics[width=5mm]{figures/mod_0.pdf}$&$\includegraphics[width=5mm]{figures/mod_0.pdf}$ \\ \midrule 

    \cite{gui2023survey} & $\includegraphics[width=5mm]{figures/mod_12.pdf}$&$\includegraphics[width=5mm]{figures/mod_0.pdf}$&$\includegraphics[width=5mm]{figures/mod_0.pdf}$&$\includegraphics[width=5mm]{figures/mod_0.pdf}$&$\includegraphics[width=5mm]{figures/mod_12.pdf}$&$\includegraphics[width=5mm]{figures/mod_0.pdf}$ \\ \midrule
    
    \cite{DBLP:journals/corr/abs-2001-00378} &$\includegraphics[width=5mm]{figures/mod_0.pdf}$&$\includegraphics[width=5mm]{figures/mod_0.pdf}$&$\includegraphics[width=5mm]{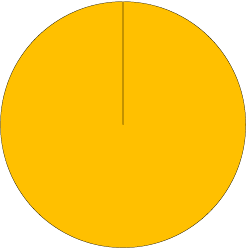}$&$\includegraphics[width=5mm]{figures/mod_0.pdf}$&$\includegraphics[width=5mm]{figures/mod_0.pdf}$&$\includegraphics[width=5mm]{figures/mod_0.pdf}$ \\ \midrule 
    
    \cite{https://doi.org/10.48550/arxiv.2203.01205} &$\includegraphics[width=5mm]{figures/mod_0.pdf}$&$\includegraphics[width=5mm]{figures/mod_0.pdf}$&$\includegraphics[width=5mm]{figures/mod_1.pdf}$&$\includegraphics[width=5mm]{figures/mod_0.pdf}$&$\includegraphics[width=5mm]{figures/mod_0.pdf}$&$\includegraphics[width=5mm]{figures/mod_0.pdf}$ \\ \midrule 
    
    \cite{9893562}&$\includegraphics[width=5mm]{figures/mod_0.pdf}$&$\includegraphics[width=5mm]{figures/mod_0.pdf}$&$\includegraphics[width=5mm]{figures/mod_1.pdf}$&$\includegraphics[width=5mm]{figures/mod_0.pdf}$&$\includegraphics[width=5mm]{figures/mod_0.pdf}$&$\includegraphics[width=5mm]{figures/mod_0.pdf}$ \\ \midrule 
    
    \cite{9770382}&$\includegraphics[width=5mm]{figures/mod_0.pdf}$&$\includegraphics[width=5mm]{figures/mod_0.pdf}$&$\includegraphics[width=5mm]{figures/mod_0.pdf}$&$\includegraphics[width=5mm]{figures/mod_1.pdf}$&$\includegraphics[width=5mm]{figures/mod_0.pdf}$&$\includegraphics[width=5mm]{figures/mod_0.pdf}$ \\ \midrule 
    
    \cite{9764632} &$\includegraphics[width=5mm]{figures/mod_0.pdf}$&$\includegraphics[width=5mm]{figures/mod_0.pdf}$&$\includegraphics[width=5mm]{figures/mod_0.pdf}$&$\includegraphics[width=5mm]{figures/mod_1.pdf}$&$\includegraphics[width=5mm]{figures/mod_0.pdf}$&$\includegraphics[width=5mm]{figures/mod_0.pdf}$ \\ \midrule 
    
    \cite{9632431}&$\includegraphics[width=5mm]{figures/mod_0.pdf}$&$\includegraphics[width=5mm]{figures/mod_0.pdf}$&$\includegraphics[width=5mm]{figures/mod_0.pdf}$&$\includegraphics[width=5mm]{figures/mod_1.pdf}$&$\includegraphics[width=5mm]{figures/mod_0.pdf}$&$\includegraphics[width=5mm]{figures/mod_0.pdf}$ \\ \midrule 
    
    \cite{DBLP:journals/corr/abs-2003-08271} &$\includegraphics[width=5mm]{figures/mod_0.pdf}$&$\includegraphics[width=5mm]{figures/mod_0.pdf}$&$\includegraphics[width=5mm]{figures/mod_0.pdf}$&$\includegraphics[width=5mm]{figures/mod_0.pdf}$&$\includegraphics[width=5mm]{figures/mod_1.pdf}$&$\includegraphics[width=5mm]{figures/mod_0.pdf}$ \\ 
    \midrule\midrule
    
    \textbf{Ours} &$\includegraphics[width=5mm]{figures/mod_0.pdf}$&$\includegraphics[width=5mm]{figures/mod_0.pdf}$&$\includegraphics[width=5mm]{figures/mod_0.pdf}$&$\includegraphics[width=5mm]{figures/mod_0.pdf}$&$\includegraphics[width=5mm]{figures/mod_0.pdf}$&$\includegraphics[width=5mm]{figures/mod_1.pdf}$ \\
    \bottomrule \bottomrule
	\end{tabular}
	    
\vspace{-10pt}
\end{table}

\subsubsection{Pretext tasks}
\label{sec-pretexttasks}
The pretext task serves as a means to learn informative representations for downstream tasks. Unlike the learning-paradigm-based criterion, the pretext-task-based criterion is also related to data modality. For example, Ericsson et al. \cite{9770283} provides a very comprehensive review of pretext tasks for multiple modalities, including image, video, text, audio, time series, and graph. The various self-supervised pretexts are divided into five broad families: transformation prediction, masked prediction, instance discrimination, clustering, and contrastive instance discrimination. Jing and Tian \cite{9086055} summarize the self-supervised feature learning methods on image and video data, and four categories are discussed: generation-based, context-based, free semantic label-based, and cross modal-based, where cross-modal-based methods construct learning task using RGB frame sequence an optical flow sequence, which are unique features in the video. Gui et al. \cite{gui2023survey} explore four kinds of pretext tasks in computer vision and natural language processing, including context-based methods, contrastive learning methods, generative algorithms, and contrastive generative methods. Essentially, the core of the pretext tasks is how to construct pseudo-supervision signals. Generally speaking, ignoring the differences in data modalities, existing pretext tasks can be roughly summarized into three categories: context prediction, instance discrimination, and instance generation. The main differences and examples are summarized in Table \ref{table-pretext}. It should be noted that here we only list some commonly used pretexts tasks, and some special pretext tasks are not the focus of this article. The details can be found in Appendix~\ref{app:pretext}.








\subsubsection{Components and modules} The literature categorizing SSCL methods according to their modules and components throughout the pipeline is also an important direction. Jaiswal et al. \cite{technologies9010002}, Le-Khac et al. \cite{9226466} and Liu et al. \cite{s23094221} sort out the modules and components required in SSL from different perspectives. Specifically, Liu et al. \cite{s23094221} summarizes the research progress of self-supervised contrastive learning on medical time series data. In summary, the pipeline can be divided into four components: positive and negative samples, pretext task, model architecture, and training loss. 

The basic intuition behind SSCL is to pull positive samples closer and push negative samples away. 
Therefore, the first component is to construct positive and negative samples. According to the suggestions of Le-Khac et al. \cite{9226466}, the main methods can be divided into the following categories: multisensory signals, data augmentation, local-global consistency, and temporal consistency. Additional descriptions regarding the characteristics of these categories can be found in Appendix~\ref{app:posneg}.

The second component is pretext tasks, which is a self-supervised task that acts as an important strategy to learn data representations using pseudo-labels \cite{technologies9010002}. Pretext tasks have been summarized and categorized in the previous subsection, so repeated content will not be introduced again. The details can be found in Section \ref{sec-pretexttasks} and Appendix~ \ref{app:pretext}.

The third component is model architecture, which determines how positive and negative samples are encoded during training. The major categories include end-to-end \cite{10.5555/3524938.3525087}, memory bank \cite{DBLP:journals/corr/abs-1912-01991}, momentum encoder \cite{9157636}, and clustering \cite{10.5555/3495724.3496555}. More details of these four architectures are summarized in Appendix~\ref{app:modelarchitecture}.

The fourth component is training loss. As summarized in \cite{9226466}, commonly used contrastive loss functions generally include scoring functions (cosine similarity), energy-based margin functions (pair loss and triplet loss), probabilistic NCE-based functions, and mutual information based functions. More details of these loss functions are summarized in Appendix~\ref{app:sslloss}.

\subsection{Surveys on time series data}
The surveys on time series data can be roughly divided into two categories. The first category focuses on different tasks, such as classification \cite{Abandasurvey,Fawzasurvey}, forecasting \cite{Lim_2021,SEZER2020106181,9461796,10.1145/3533382}, and anomaly detection \cite{10.1145/3444690,8926446}. These surveys comprehensively sort out the existing methods for each task. The second category focuses on the key components of time series modeling based on deep neural networks, such as data augmentation \cite{ijcai2021p631,DBLP:journals/corr/abs-2007-15951,ncaaug2023,s23094221}, model structure \cite{wen2022tstransformers,10.1145/3559540,s23094221}. \cite{ijcai2021p631} proposed a new taxonomy that divides the existing data augmentation techniques into basic and advanced approaches. \cite{DBLP:journals/corr/abs-2007-15951} also provides a taxonomy and outlines four families: transformation-based methods, pattern mixing, generative models, and decomposition methods. Moreover, both \cite{ijcai2021p631} and \cite{DBLP:journals/corr/abs-2007-15951} empirically compare different data augmentation methods for time series classification tasks. \cite{wen2022tstransformers} systematically reviews transformer schemes for time series modeling from two perspectives: network structure and applications. Liu et al. \cite{s23094221} provide a comprehensive summary of the various augmentations applied to medical time series data, the architectures of pre-training encoders, the types of fine-tuning classifiers and clusters, and the popular contrastive loss functions. The taxonomies proposed by Eldele et al. \cite{eldele2023labelefficient}, Deldari et al. \cite{DBLP:journals/corr/abs-2206-02353} and Liu et al. \cite{s23094221} are somewhat similar to our proposed taxonomy, i.e., three taxonomies involve time series self-supervised contrastive learning methods. However, our taxonomy provides more detailed categories and more literature in the contrastive-based approach. Although the taxonomy proposed by Liu et al.
\cite{s23094221} also focuses on time series data. They emphasize discussion of medical time series data, while we focus more on general time series SSL. More importantly, in addition to contrastive-based approaches, we also thoroughly review a large set of literature for the generative-based and adversarial-based approaches.

\section{Generative-based Methods}
\label{sec:generative}

In this category, the pretext task is to generate the expected data based on a given view of the data. In the context of time series modeling, the commonly used pretext tasks include using the past series to forecast the future windows or specific time stamps, using the encoder and decoder to reconstruct the input, and forecasting the unseen part of the masked time series. This section sorts out the existing self-supervised representation learning methods in time series modeling from the perspectives of autoregressive-based forecasting, autoencoder-based reconstruction, and diffusion-based generation. It should be noted that the autoencoder-based reconstruction task is also viewed as an unsupervised framework. In the context of SSL, we mainly use the reconstruction task as a pretext task, and the final goal is to obtain the representations through autoencoder models. The illustration of the generative-based SSL for time series is shown in Fig. \ref{fig:gen_ae}. In Appendix~\ref{app:arf} - \ref{app:diff}, the main advantages and disadvantages of three generative-based submethods are summarized. Furthermore, the direct comparison of the three methods is shown in Appendix~\ref{app:gensummary}. 


\subsection{Autoregressive-based forecasting}

Given the current time step $t$, the goal of an autoregressive-based forecasting (ARF) task is to forecast $K$ future horizons based on $t$ historical time steps, which can be expressed as:
\begin{equation}
	\hat{x}_{[t+1:t+K]} = f(x_{[1:t]}),
	\label{eq:ar}
\end{equation}
where $\hat{x}_{[t+1:t+K]}$ represents the target window, and $K$ represents the length of the target window. When $K=1$, (\ref{eq:ar}) is a single-step forecasting model, and it is a multi-step forecasting model when $K > 1$. $x_{[1:t]}$ represents the input series before time $t$ (including $t$), which is usually used as the input of the model. $f(\cdot)$ represents the forecasting model. The learning objective is to minimize the distance between the predicted target window and the ground truth, thus the loss function can be defined as:
\begin{equation}
	\mathcal{L} = D(\hat{x}_{[t+1:t+K]}, x_{[t+1:t+K]}),
	\label{eq:arloss}
\end{equation}
where $D(\cdot)$ represents the distance between the predicted future window $\hat{x}_{[t+1:t+K]}$ and the ground-truth future window $x_{[t+1:t+K]}$, usually measured by the mean square error (MSE), i.e.,
\begin{equation}
	\mathcal{L} = \frac{1}{K}\sum_{k=1}^{K}(\hat{x}_{[t+k]} - x_{[t+k]})^2.
	\label{eq:mse}
\end{equation}

    



In the time series modeling with autoregressive-based forecasting task as a pretext task, Recurrent neural networks (RNNs) are widely used thanks to their strong capability in spatiotemporal dynamic behavior modeling or sequence prediction \cite{9461796,10.1145/3533382,pmlr-v162-schirmer22a,Tan_Ye_Yang_Liu_Ma_Yip_Wong_Yuen_2020}. Therefore, it is also naturally applied in the pretext task based on autoregressive forecasting. THOC \cite{NEURIPS2020_97e401a0} constructs a self-supervised pretext task for multi-resolution single-step forecasting called Temporal Self-Supervision (TSS). TSS takes the L-layer dilated RNN with skip-connection structure as the model. By setting skip length, it can ensure that the forecasting tasks can be performed with different resolutions at the same time. In addition to RNNs, the forecasting models based on Convolutional neural networks (CNNs) also have been developed \cite{10.1007/978-3-030-47426-3_39}. Moreover, STraTS \cite{10.1145/3516367} first encodes the time series data into triple representations to avoid the limitations of using basic RNN and CNN in modeling irregular and sparse time series data and then builds the transformer-based forecasting model for modeling multivariate medical clinical time series. Graph-based time series forecasting methods can also be used as a self-supervised pretext task. Compared with RNNs and CNNs, Graph Neural Networks (GNNs) can better capture the correlation among variables and constituent in multivariate time series data, such as GDN \cite{Deng_Hooi_2021} and GTS \cite{shang2021discrete}. Graph-augmented normalizing flow (GANF) is another graph-based approach that can model the conditional dependencies among constituent time series \cite{dai2022graphaugmented}. In order to choose a more appropriate model in building time series SSL task, we further give the advantages and disadvantages of these three commonly used models. The details can be found in Appendix~\ref{app:rnncnngnn}. Unlike the above methods, SSTSC \cite{XI2022105331} proposes a temporal relation learning prediction task based on the ``Past-Anchor-Future" strategy as a self-supervised pretext task. Instead of directly forecasting the values of the future time windows, SSTSC predicts the relationships of the time windows, which can fully mine the temporal relationship in the data.

\begin{figure}[!t]
	\centering 
	\subfloat[Autoregressive-based forecasting task]{\includegraphics[width=75mm]{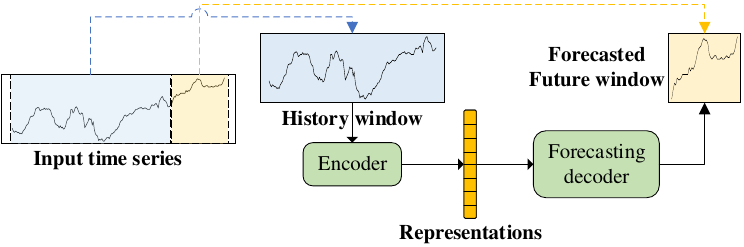}}\\
	\subfloat[Autoencoder-based reconstruction task]{\includegraphics[width=75mm]{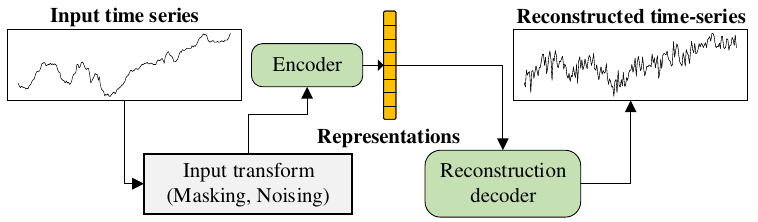}}\\
	\subfloat[Diffusion-based generation task]{\includegraphics[width=75mm]{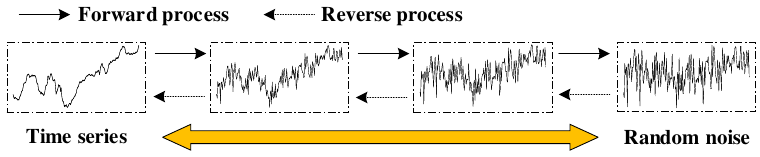}}\\
	\caption{Three categories of generative-based SSL for time series data.}
 \label{fig:gen_ae}
\vspace{-10pt}
\end{figure}

\subsection{Autoencoder-based reconstruction}
The autoencoder is an unsupervised artificial neural network composed of an encoder and a decoder \cite{pmlr-v27-baldi12a}. The encoder maps the input $x$ to the representation $z$, and then the decoder re-maps the representation $z$ back to the input. The output of the decoder is defined as the reconstructed input $\tilde{x}$. The process can be expressed as:
\begin{equation}\label{eq:encoder}
	z = E(x), \quad \tilde{x} = D(z),
\end{equation}
where $E(\cdot)$ and $D(\cdot)$ represent the encoder and decoder, respectively. The difference between the original input $x$ and the reconstructed input $\tilde{x}$ is called the reconstruction error, and the goal of the self-supervised pretext task using autoencoder structure is to minimize the error between $x$ and $\tilde{x}$, i.e.,
\begin{equation}
	\mathcal{L} = \left \| x-\tilde{x} \right \|_2.
	\label{eq:aemse}
\end{equation}

The model structure of (\ref{eq:encoder}) is defined as the basic autoencoder (BAE). Most BAE-based methods jointly train the encoder $E(\cdot)$ and the decoder $D(\cdot)$. Then removing the decoder $D(\cdot)$ and leaving only the encoder $E(\cdot)$ that is used as a feature extractor, and the representation $z$ is used for downstream tasks \cite{10.5555/3327546.3327715,DBLP:journals/corr/MalhotraTVAS17,PTLSTMSAESR,10.1145/3447548.3467174}. For example, TimeNet \cite{DBLP:journals/corr/MalhotraTVAS17}, PT-LSTM-SAE\cite{PTLSTMSAESR}, and Autowarp \cite{10.5555/3327546.3327715} all use RNN to build a sequence autoencoder model including encoder and decoder, which tries to reconstruct the input series. Once the model is learned, the encoder is used as a feature extractor to obtain an embedded representation of time series samples, which can help downstream tasks, such as classification and forecasting, achieve better performance. Zhang et al. \cite{9455535} build a CNN-based autoencoder model and keep the encoder as a feature extractor after minimizing (\ref{eq:aemse}). The experimental results show that using the encoded representation is better than directly using the original time series data in industrial fault detection tasks.

However, the representations obtained by (\ref{eq:aemse}) are sometimes task-agnostic. Therefore, it is feasible to introduce additional training constraints based on (\ref{eq:aemse}). Abdulaal et al. \cite{10.1145/3447548.3467174} focus on the complex asynchronous multivariate time series data and introduce the spectral analysis in the autoencoder model. The synchronous representation of the time series is extracted by learning the phase information in the data, which is eventually used for the anomaly detection task. DTCR \cite{NEURIPS2019_1359aa93} is a temporal clustering-friendly representation learning model. It introduces K-means constraints in the reconstruction task, making the learned representation more friendly to clustering tasks. USAD \cite{10.1145/3394486.3403392} uses an encoder and two decoders to build an autoencoder model and introduces adversarial training based on (\ref{eq:aemse}) to enhance the representation ability of the model. FuSAGNet \cite{10.1145/3534678.3539117} introduces graph learning on the sparse autoencoder to model relationships in multivariate time series explicitly. 

Denoising autoencoder (DAE) is another widely used approach, which is based on the addition of noise to the input series to corrupt the data, and then followed by the reconstruction task \cite{10.1145/1390156.1390294}. DAE can be formulated as:
\begin{equation}
	x_n = \mathcal{T}(x),\quad Z = E(x_n), \quad \tilde{x} = D(z),
\end{equation}
where $\mathcal{T}$ indicates the operation that adds noise. The learning objective of a DAE is the same as that of a BAE, which is to minimize the difference between $x$ and $\tilde{x}$. In time series modeling, more than one method can add noise to the input, such as adding Gaussian noise \cite{8059861,8982996} and randomly setting some time steps to zero \cite{ZHENG2022159, LI201523}.

Mask autoencoder (MAE) is a structure widely used in language models and vision models in recent years \cite{devlin-etal-2019-bert,9879206}. The core idea behind MAE is that in the pre-training phase, the model first masks part of the input and then tries to predict the masked part through the unmasked part. Unlike BAE and DAE, the loss of MAE is only computed on the masked part. MAE can be formulated as:
\begin{equation}
	x_{m} = \mathcal{M}(x), \quad z = E(x_{m}), \quad \tilde{x} = D(z),
\end{equation}
\begin{equation}
	\mathcal{L} = \mathcal{M} (\left \| x-\tilde{x} \right \|_2),
	\label{eq:maskaeloss}
\end{equation}
where $\mathcal{M}(\cdot)$ represents the mask operation, and $X_{m}$ represents the masked input. In language models, since the input is usually a sentence, the mask operation masks some words in a sentence or replaces them with other words. In vision models, the mask operation will mask the pixels or patches in an image. For time series data, a feasible operation is to mask part of the time steps and then use the unmasked part to predict the masked time steps. Existing masking methods for time series data can be divided into three categories: time-step-wise masking, segment-wise masking, and variable-wise masking.  

The time-step-wise masking randomly selects a certain proportion of time-steps in the series to mask, so the fine-grained information is easier to capture, but it is difficult to learn contextual semantic information in time series. The segment-wise masking randomly selects segments to mask, which allows the model to pay more attention to slow features in the time series, such as trends or high-level semantic information. STEP \cite{10.1145/3534678.3539396} divided the series into multiple non-overlapping segments of equal length and then randomly selected a certain proportion of the segments for masking. Moreover, STEP pointed out two advantages of using segment-wise masking: the ability to capture semantic information and reduce the input length to the encoder. Different from STEP, Zerveas et al. \cite{10.1145/3447548.3467401} performed a more complex masking operation on the time series, i.e., the multivariate time series was randomly divided into multiple non-overlapping segments of unequal length on each variable. Variable-wise masking was introduced by Chauhan et al.\cite{10.1145/3534678.3539394}, who defined a new time series forecasting task called variable subset forecast (VSF). In VSF, the time series samples used for training and inference have different dimensions or variables, which may be caused by the absence of some sensor data. This new forecasting task brings the feasibility of self-supervised learning based on variable-wise masking. Unlike random masking, TARNet \cite{10.1145/3534678.3539329} considers the pre-trained model based on the masking strategy irrelevant to the downstream task, which leads to sub-optimal representations. TARNet uses self-attention score distribution from downstream task training to determine the time steps that require masking.


Variational autoencoder (VAE) is a model based on variational inference \cite{https://doi.org/10.48550/arxiv.1312.6114,DBLP:journals/corr/abs-1906-02691}. The encoder encodes the input $x$ to the probability distribution $P(z|x)$ instead of the explicit representation $z$. When the decoder is used to reconstruct the input, a vector generated by sampling from the distribution $P(z|x)$ will be used as input to the decoder. The process can be expressed as:
\begin{equation}
	P(z|x) = E(x), \quad z = \mathcal{S}(P(z|x)), \quad \tilde{x} = D(z),
\end{equation}
where $\mathcal{S}(\cdot)$ represents the sampling operation. Unlike (\ref{eq:aemse}), the loss function of a VAE includes two terms: the reconstruction item and the regularization item, i.e.,
\begin{equation}
	\mathcal{L} = \left \| x-\tilde{x} \right \|_2 + \text{KL}(\mathcal{N}(\mu,\delta), \mathcal{N}(0,I)),
	\label{eq:vaeloss}
\end{equation}
where $\text{KL}(\cdot)$ represents the Kullback-Leibler divergence. The role of the regularization term is to ensure that the learned distribution $P(z|x)$ is close to the standard normal distribution, thereby regulating the representation of the latent space. The representation learning method based on VAE can model the distribution of each time step to better capture the complex spatiotemporal dependencies and provide better interpretability in time series modeling tasks. For example, InterFusion \cite{10.1145/3447548.3467075} is a hierarchical VAE that models inter-variable and temporal dependencies in time series data. OmniAnomaly  \cite{10.1145/3292500.3330672} combines VAE and Planar Normalizing Flow to propose an interpretable time series anomaly detection algorithm. In order to better capture the dependencies between different variables in multivariate time series, GRELEN \cite{ijcai2022p332} and VGCRN \cite{pmlr-v162-chen22x} introduce the graph structure and in VAE. In addition to modeling on regular time series, the methods based on VAE have made progress in sparse and irregular time series data representation learning, such as mTANs \cite{shukla2021multitime}, P-VAE \cite{pmlr-v119-li20k} and HetVAE \cite{DBLP:journals/corr/abs-2107-11350}. The latest work attempts to extract seasonal and trend representations in time series data based on VAE. LaST \cite{wang2022learning} is a disentangled variational inference framework with mutual information constraints. It separates seasonal and trend representations in the latent space to achieve accurate time series forecasting.



\subsection{Diffusion-based generation}
As a new kind of deep generative model, diffusion models have achieved great success recently in many fields, including image synthesis, video generation, speech generation, bioinformatics, and natural language processing due to their powerful generating ability~\cite{ho2020denoising,dhariwal2021diffusion,yang2022diffusion,cao2022survey,luo2022understanding}. The key design of the diffusion model contains two inverse processes: the forward process of injecting random noise to destruct data and the reverse process of sample generation from noise distribution (usually normal distribution). The intuition is that if the forward process is done step-by-step with a transition kernel between any two adjacent states, then the reverse process can follow a reverse state transition operation to generate samples from noise (the final state of the forward process). However, it is usually not easy to formulate the reverse transition kernel, and thus diffusion models learn to approximate the kernel by deep neural networks. Nowadays, there are mainly three basic formulations of diffusion models: denoising diffusion probabilistic models (DDPMs)~\cite{ho2020denoising,sohl2015deep}, score matching diffusion models~\cite{song2019generative,song2020improved}, and score SDEs~\cite{song2021maximum,song2020score}. 


For DDPMs, the forward and reverse processes are two Markov chains: a forward chain that adds random noise to data and a reverse chain that transforms noise back into data. Formally, denote the data distribution as $\boldsymbol{x}_0 \sim q(\boldsymbol{x}_0)$, the forward Markov process gradually adds Gaussian noise to the data according to transition kernel $q(\boldsymbol{x}_t|\boldsymbol{x}_{t-1})$. It generates a sequence of random variables $\boldsymbol{x}_1, \boldsymbol{x}_2, \dots, \boldsymbol{x}_T$. Thus the joint distribution of $\boldsymbol{x}_1, \boldsymbol{x}_2, \dots, \boldsymbol{x}_T$ conditioned on $\boldsymbol{x}_0$ is 
\begin{equation}
q(\boldsymbol{x}_1, \boldsymbol{x}_2, \dots, \boldsymbol{x}_T|\boldsymbol{x}_0) = \prod_{t=1}^{T} q(\boldsymbol{x}_t | \boldsymbol{x}_{t-1}).
\end{equation}
For simplicity of calculation, the transition kernel is usually set as 
\begin{equation}
q(\boldsymbol{x}_t|\boldsymbol{x}_{t-1}) = \mathcal{N}(\boldsymbol{x}_t;\sqrt{1-\beta_{t}}\boldsymbol{x}_{t-1}, \beta_t I),
\end{equation}
where $\beta_1, \beta_2, \dots, \beta_T$ is a variance schedule of the forward process (usually chosen $\beta_t \in (0,1)$ ahead of model training) and $p(\boldsymbol{x}_T)=\mathcal{N}(\boldsymbol{x}_T; 0, I).$ 
Similarly, the joint distribution of the reverse process is 
\begin{equation}
p_{\theta}(\boldsymbol{x}_0, \boldsymbol{x}_1, \dots, \boldsymbol{x}_T) = p(\boldsymbol{x}_T) \prod_{t=1}^{T}p_{\theta}(x_{t-1}|x_t),
\end{equation}
where $\theta$ is the model parameters and $p_{\theta}(\boldsymbol{x_{t-1}|x_t}) = \mathcal{N}(\boldsymbol{x}_{t-1}; \mu_{\theta}(\boldsymbol{x}_t, t), \sum_{\theta}(\boldsymbol{x}_t,t)).$ 
The key to achieving the success of sample generating is training the parameters $\theta$ to match the actual reverse process, that is, minimizing the Kullback-Leibler divergence between the two joint distributions. Thus, according to Jensen's inequality, the training loss is 
\begin{equation}
\begin{aligned}
\textbf{KL}(q(\boldsymbol{x}_1, \boldsymbol{x}_2, \dots, \boldsymbol{x}_T)||p_{\theta}(\boldsymbol{x}_0, \boldsymbol{x}_1, \dots, \boldsymbol{x}_T)) \\
\geq \mathbf{E}[-\log p_{\theta}(\boldsymbol{x}_0)] + const.
\end{aligned}
\end{equation}
 



For score-based diffusion models, the key idea is to perturb data with a sequence of Gaussian noise and then jointly estimate the score functions for all noisy data distributions by training a deep neural network conditioned on noise levels. The motivation of the idea is that, in many situations, it is easier to model and estimate the score function than the original probability density function.
Langevin dynamics is one of the proper techniques. With a step size $\alpha>0$, the number of iterations $T$, and an initial sample $x_0$, Langevin dynamics iteratively does the following estimation to gain a close approximation of $p(\boldsymbol{x})$
\begin{equation}
\boldsymbol{x}_t \leftarrow \boldsymbol{x}_{t-1} + \alpha \nabla_x \log p(\boldsymbol{x}_{t-1}) + \sqrt{2\alpha} \boldsymbol{z}_t, 1\leq t \leq T  ,
\end{equation}
where $\boldsymbol{z}_t \sim \mathcal{N}(0, I)$. However, the score function is inaccurate without the training data, and Langevin dynamics may not converge correctly. Thus, the key approach (NCSN, a noise-conditional score network), perturbing data with a noise sequence and jointly estimating the score function for all the noisy data with a deep neural network conditioned on noise levels, is proposed~\cite{song2019generative}. Training and sampling are decoupled in score-based generative models, which inspires different choices in such two processes~\cite{song2020improved}. 

For score SDEs, the diffusion operation is processed according to the stochastic differential equation (SDE)~\cite{song2020score}:
\begin{equation}\label{SDE}
    d\boldsymbol{x} = f(\boldsymbol{x}, t)dt + g(t)d\boldsymbol{w},
\end{equation}
where $f(\boldsymbol{x}, t)$ and $g(t)$ are diffusion function and drift function of the SDE, respectively, and $\boldsymbol{w}$ is a standard Wiener process. Different from DDPMs and SGMs, Score SDEs generalize the diffusion process to the case of infinite time steps. 
Fortunately, DDPMs and SGMs also can be formulated with corresponding SDEs. For DDPMs, the SDE is
\begin{equation}
    d\boldsymbol{x} = -\frac{1}{2}\beta(t)\boldsymbol{x}dt + \sqrt{\beta(t)}d\boldsymbol{w},
\end{equation}
where $\beta(\frac{t}{T}) = T\beta_t$ when T goes to infinity; for SGMs, the SDE is
\begin{equation}
    d\boldsymbol{x} = \sqrt{\frac{d[\delta(t)^2]}{dt}}d\boldsymbol{w},
\end{equation}
where $\delta(\frac{t}{T})=\delta_t$ as T goes to infinity. 
With any diffusion process in the form of (\ref{SDE}), the reverse process can be gained by solving the following SDE:
\begin{equation}
    d\boldsymbol{x} = [f(\boldsymbol{x}, t) - g(t)^2 \nabla_{\boldsymbol{x}} \log q_t(\boldsymbol{x})]dt + g(t)d{\boldsymbol{\overline w}},
\end{equation}
where $\boldsymbol{\overline{w}}$ is a standard Wiener process when time flows reversely and $dt$ is an infinitesimal time step. Besides that, the existence of an ordinary differential equation, which is also called the \textit{probability flow ODE}, is defined as follows. 
\begin{equation}
 d\boldsymbol{x} = [f(\boldsymbol{x}, t)-{\frac{1}{2}}{g(t)}^2\nabla_{\boldsymbol{x}}\log q_t(\boldsymbol{x})]dt.
\end{equation}
The trajectories of the probability flow ODE have the same marginals as the reverse-time SDE. Once the score function at each time step is known, the reverse SDE can be solved with various numerical techniques. Similar objective is designed with SGMs. 

Diffusion models have also been applied in time series analysis recently. We briefly summarize them based on the designed architectures and the main diffusion techniques used. Conditional score-based diffusion models for imputation (CSDI)~\cite{tashiro2021csdi} were proposed for time series imputation task. CSDI utilizes score-based diffusion models conditioned on observed data. In time series forecasting tasks, TimeGrad~\cite{rasul2021autoregressive} takes an RNN conditioned diffusion probabilistic model at some time step to depict the fixed forward process and the learned reverse process. D$^3$VAE~\cite{li2023generative} is a bidirectional variational auto-encoder (BVAE) equipped with diffusion, denoise, and disentanglement. In D$^3$VAE, the coupled diffusion process augments the input time series and output time series simultaneously. 
ImDiffusion~\cite{chen2023imdiffusion} combines imputation and diffusion models for time series anomaly detection. SSSD~\cite{alcaraz2022diffusion} combines diffusion models and structured state space models for time series imputation and forecasting tasks. 
DiffLoad~\cite{wang2023diffload} proposes a diffusion-based structure for electrical load probabilistic forecasting by considering both epistemic and aleatoric uncertainties. DiffSTG~\cite{wen2023diffstg} presents the first shot to predict the evolution of spatio-temporal graphs using DDPMs. 

\begin{figure*}[!t]
	\centering
	\subfloat[Sampling contrast]{\includegraphics[width=35mm]{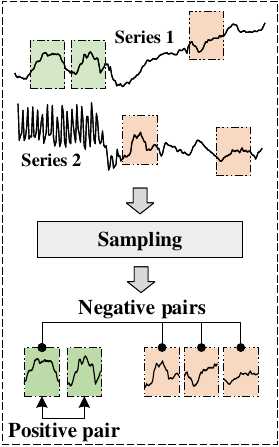}}
	\subfloat[Prediction contrast]{\includegraphics[width=35mm]{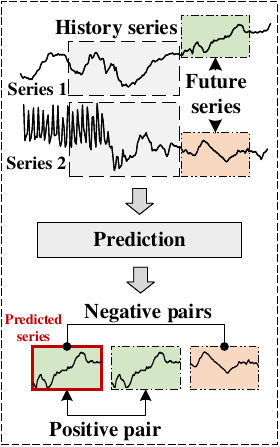}}
	\subfloat[Augmentation contrast]{\includegraphics[width=35mm]{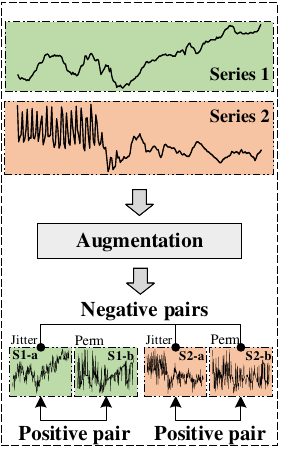}}
	\subfloat[Prototype contrast]{\includegraphics[width=35mm]{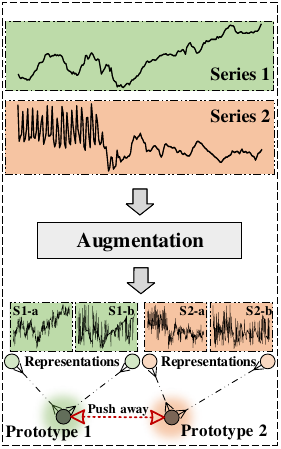}}
	\subfloat[Expert knowledge contrast]{\includegraphics[width=35mm]{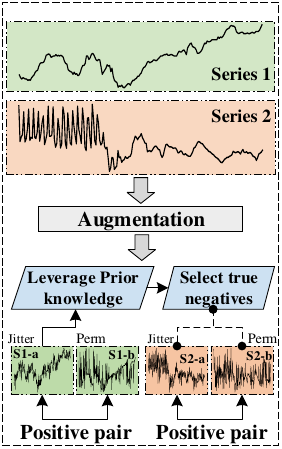}}
	\caption{Five categories of contrastive-based SSL for time series data.}
	\label{fig:contrastshow}
    \vspace{-10pt}
\end{figure*}

\section{Contrastive-based Methods}
\label{sec:contrastive}

Contrastive learning is a widely used self-supervised learning strategy, showing a strong learning ability in computer vision and natural language processing. Unlike discriminative models that learn a mapping rule to true labels and generative models that try to reconstruct inputs, contrastive-based methods aim to learn data representations by contrasting between positive and negative samples. Specifically, positive samples should have similar representations, while negative samples have different representations. Therefore, the selection of positive samples and negative samples is very important to contrastive-based methods. This section sorts out and summarizes the existing contrastive-based methods in time series modeling according to the selection of positive and negative samples. The illustration of the contrastive-based SSL for time series is shown in Fig. \ref{fig:contrastshow}. In Appendix~\ref{app:sc} - \ref{app:ekc}, the main advantages and disadvantages of five contrastive-based submethods are summarized.


\subsection{Sampling contrast}
Sampling contrast follows a widely used assumption in time series analysis that two neighboring time windows or time stamps have a high degree of similarity, so positive and negative samples are directly sampled from the raw time series, as shown in Fig. \ref{fig:contrastshow}(a). Specifically, given a time window (or a timestamp) as an anchor, its nearby window (or the time stamp) is more likely to be similar (small distance), and the distant window (or the time stamp) should be less similar (large distance). The term ``similar" indicates that two windows (or two-time stamps) have more common patterns, such as the same amplitude, the same periodicity, and the same trend. 

As mentioned in \cite{NEURIPS2019_53c6de78}, suppose one anchor $x^{ref}$, one positive sample $x^{pos}$, and $K$ negative samples ${x^{neg}_k}_{, \ k \in 1,2,\cdots,K}$ are chosen, we expect to assimilate $x^{ref}$ and $x^{pos}$ and to distinguish between $x^{ref}$ and $x^{neg}_k$, i.e.,
\begin{equation}
    \mathcal{L} = -\log(S(x^{ref}, x^{pos})) - \sum_{k=1}^K \log(-S(x^{ref}, x^{neg}_k)),
    \label{eq:cs1}
\end{equation}
where $S(\cdot)$ denotes the similarity of the two representations. However, due to the non-stationary characteristics of most time series data, it is still a challenge to choose the correct positive and negative samples based on contextual information in time series data. Temporal neighborhood coding (TNC) was recently proposed to deal with this problem \cite{DBLP:journals/corr/abs-2106-00750}. TNC uses augmented Dickey-Fuller (ADF) statistical test to determine the stationary region and introduces positive-unlabeled (PU) learning to handle the problem of sampling bias by treating negative samples as unknown samples and then assigning weights to these samples. The learning objective is extended to
\begin{equation}
    \begin{aligned}
    \mathcal{L} = 
    & -\mathbb{E}_{x^{pos}[ \in \mathcal{N}} \log S(x^{ref},x^{pos}) ]\\ 
    & -  \mathbb{E}_{x^{neg} \in \Tilde{\mathcal{N}}} [(1-w) \times \log -S(x^{ref},x^{neg}) \\
    & +  w \times \log S(x^{ref},x^{neg})],
    \end{aligned}
\end{equation}
where $w$ is the probability of sampling false negative samples, $\mathcal{N}$ denotes the neighboring area, and $\Tilde{\mathcal{N}}$ denotes the non-neighboring area. Supervised contrastive learning (SCL) \cite{NEURIPS2020_d89a66c7} effectively addresses sampling bias, so introducing the supervised signal to identify positive and negative samples is a feasible solution. Neighborhood contrastive learning (NCL) is a recent time series modeling method that combines context sampling and the supervised signal to generate positive and negative samples \cite{pmlr-v139-yeche21a}. NCL assumes that if two samples share some predefined attributes, then they are considered to share the same neighboring area.



\subsection{Prediction contrast}
In this category, prediction tasks that use the context (present) to predict the target (future information) are considered self-supervised pretext tasks, and the goal is to maximally preserve the mutual information of the context and the target. Contrastive predictive coding (CPC) proposed by \cite{DBLP:journals/corr/abs-1807-03748} provides a contrastive learning framework to perform the prediction task using InfoNCE loss. As shown in Fig.~\ref{fig:contrastshow}(b), the context $c_t$ and the sample from $p(x_{t+k}|c_t)$ constructs positive pairs, and the samples from the `proposal’ distribution $p(x_{t+k})$ are negative samples. The learning objective is as follows:
\begin{equation}
    \mathcal{L}=-\underset{X}{\mathbb{E}}\left[\log \frac{f_{k}\left(x_{t+k}, c_{t}\right)}{\sum_{x_{j} \in X} f_{k}\left(x_{j}, c_{t}\right)}\right],
    \label{eq:cpc}
\end{equation}
where $f_k(\cdot)$ is the density ratio that preserves the mutual information of $c_t$ and $x_{t+k}$ \cite{DBLP:journals/corr/abs-1807-03748}, and it can be estimated by a simple log-bilinear model:
\begin{equation}
    f_{k}\left(x_{t+k}, c_{t}\right) = \exp (z^T_{t+k}W_kc_t).
    \label{eq:cpc2}
\end{equation}

It can be seen that CPC does not directly predict future observations $x_{t+k}$. Instead, it tries to preserve the mutual information of $c_t$ and $x_{t+k}$. This allows the model to capture the ``slow features" that span multiple time steps. Following the architecture of CPC, LNT\cite{DBLP:journals/corr/abs-2202-03944}, TRL-CPC\cite{DBLP:journals/corr/abs-2202-03639}, TS-CP$^2$\cite{10.1145/3442381.3449903}, and Skip-Step CPC\cite{kexinnipsws2022} were proposed. LNT and TRL-CPC use the same structure as the original CPC \cite{DBLP:journals/corr/abs-1807-03748} to build a representation learning model, and the purpose is to capture the local semantics across the time to detect the anomaly points. TS-CP$^2$ and Skip-Step CPC replace the autoregressive model in the original CPC structure with TCN \cite{DBLP:journals/corr/abs-1803-01271}, which improves feature learning ability and computational efficiency. Moreover, Skip-Step CPC points out that adjusting the distance between context representation $c_t$ and $x_{t+k}$ can construct different positive pairs, which leads to different results in time series anomaly detection.  

In addition to the basic contextual prediction tasks mentioned before, some more complex prediction tasks were constructed and proved useful. CMLF \cite{10.1145/3459637.3482483} transforms time series into coarse-grained and fine-grained representations and proposes a multi-granularity prediction task. This allows the model to represent the time series at different scales. TS-TCC\cite{ijcai2021-324} and its extended version CA-TCC\cite{emadeldeen2022catcc} designed a cross prediction task, which uses the context of $x^{T1}$ to predict the target in $x^{T2}$, and vice versa uses the context of $x^{T2}$ to predict the target in $x^{T1}$.


 

\subsection{Augmentation contrast}

Augmentation contrast is one of the most widely used contrastive frameworks, as shown in Fig. \ref{fig:contrastshow}(c). Most methods utilize data augmentation techniques to generate different views of an input sample and then learn representations by maximizing the similarity of the views that come from the same sample and minimizing the similarity of the views that come from the different samples. SimCLR \cite{10.5555/3524938.3525087} is a very typical multi-view invariance-based representation learning framework, which has been used in many subsequent methods. The objective function based on this framework is:
\begin{equation}
	\label{eq:simclr}
    \mathcal{L}=-\log \frac{\exp \left(\operatorname{sim}\left(\boldsymbol{z}_{1}, \boldsymbol{z}_{2}\right) / \tau\right)}
    {\sum_{k=1}^{2N} \mathbbm{1}_{[k \neq 1]}  \exp \left(\operatorname{sim}\left(\boldsymbol{z}_{1}, \boldsymbol{z}_{k}\right) / \tau\right)},
\end{equation}
where $\tau$ is temperature parameter, $\operatorname{sim}(\cdot)$ represents the similarity between two representation vectors, and $z_k$ represents the training samples in a batch. It can be considered that in the feature learning framework based on multi-view invariance, the core is to obtain different views of the input samples. When handling images in computer vision, commonly used data augmentation methods include cropping, scaling, adding noise, rotation, and resizing \cite{10.5555/3524938.3525087}. However, compared with augmentation methods for images, the augmentation methods for time series needs to consider both temporal and variable dependencies.

Since time series data can be converted to frequency domain representations through Fourier transform, the augmentation method can be developed from the time and frequency domains. In the time domain, TS-TCC\cite{ijcai2021-324} and its extended version CA-TCC\cite{emadeldeen2022catcc} designed two time series data augmentation techniques, one is strong augmentation (permutation-and-jitter), and the other is weak augmentation (jitter-and-scale). TS2Vec \cite{Yue_Wang_Duan_Yang_Huang_Tong_Xu_2022} generates different views through masking operations that randomly mask out some time steps.  Generally speaking, there is no one-size-fits-all answer to the choice of data augmentation methods. Therefore, some works comprehensively compare and study the augmentation methods and further evaluate the performance on different tasks \cite{POPPELBAUM2022108397,ijcai2021p631,9539005,Iwana_2021}. All the above methods only need a single time series sample in the augmentation operation, while Mixing-up \cite{WICKSTROM202254} fuses two time series samples to generate a newly augmented view, while the pretext task is to correctly predict the proportion of two original time series samples in augmented view.  


Data augmentation in the frequency domain is also feasible for time series data. CoST \cite{woo2022cost} is a disentangled seasonal-trend representation learning method, which uses fast Fourier transform to convert different augmented views into amplitude and phase representations, and then uses (\ref{eq:simclr}) to train the model. BTSF \cite{pmlr-v162-yang22e} is a contrastive-based method based on a time-frequency fusion strategy, which first generates an augmented view in the time domain through the dropout operation and then generates another augmented view in the frequency domain through Fourier transform. Finally, the bilinear temporal-spectral fusion mechanism is used to achieve the fusion of time-frequency information. However, CoST and BTSF do not modify the frequency representation, while TF-C \cite{nips-tfc} directly augments the time series data through frequency perturbations, which has achieved better performance than TS2Vec~\cite{Yue_Wang_Duan_Yang_Huang_Tong_Xu_2022} and TS-TCC~\cite{ijcai2021-324}. Specifically, TF-C implements three augmentation strategies: low- vs. high-band perturbations, single- vs. multi-component perturbations, and random vs. distributional perturbations. 



In addition to the above methods, many view generation methods are closely related to downstream tasks. Recently, DCdetector~\cite{DCdetector2023} proposes a dual attention contrastive representation framework for time series anomaly detection. The in-patch and patch-wise representations are designed to gain two views of the input samples, as normal samples behave differently from abnormal ones in such two views. TimeCLR~\cite{YANG2022108606} proposed DTW augmentation, which can not only simulate phase shifts and amplitude changes but also retain the structure and characteristics of the time series. CLOCS \cite{pmlr-v139-kiyasseh21a} is a self-supervised pre-training method for medical and physiological signals, which uses multi-view invariance contrast in the three perspectives of time, space, and patient to promote higher similarity of representations from the same source. CLUDA \cite{ozyurt2023contrastive} introduces multi-view invariance contrast in the time series domain adaptation problem, which captures the contextual representation of time series data through intra-domain and inter-domain contrast. MTFCC~\cite{10001758} is another view generation method based on multi-scale characteristics, which samples time series samples at multiple scales and considers that the views from the same sample have similar representations, even if their scales are different. Methods for constructing multiple contrastive views based on multi-granularity or multi-scale augmentations also include MRLF\cite{10.1145/3485447.3512056}, CMLF \cite{10.1145/3459637.3482483}, and SSLAPP\cite{ijcai2022p537}.


\subsection{Prototype contrast}

The contrastive learning framework based on (\ref{eq:cpc}) and (\ref{eq:simclr}) is essentially an instance discrimination task, which encourages samples to form a uniform distribution in the feature space \cite{10.5555/3524938.3525859}. However, the real data distribution should satisfy that the samples of the same class are more concentrated in a cluster, while the distance between different clusters should be farther. SCL \cite{NEURIPS2020_d89a66c7} is an ideal solution when real labels are available, but this is difficult to implement in practice, especially for time series data. Therefore, introducing clustering constraints into existing contrastive learning frameworks is an alternative, such as CC \cite{Li_Hu_Liu_Peng_Zhou_Peng_2021}, PCL \cite{li2021prototypical}, and SwAV \cite{10.5555/3495724.3496555}. PCL and SwAV 
contrast the samples with the constructed prototypes, i.e., the cluster centers, which reduces the computation and encourages the samples to present a cluster-friendly distribution in the feature space. An illustration of prototype contrast is shown in Fig. \ref{fig:contrastshow}(d).

In time series modeling based on prototypes contrast, ShapeNet \cite{Li_Choi_Xu_S_Bhowmick_Chun_Wong_2021} takes shapelets as input and constructs a cluster-level triplet loss, which considers the distance between the anchor and multiple positive (negative) samples as well as the distance between positive (negative) samples. ShapeNet is an implicit prototype contrast because it does not introduce explicit prototypes (cluster centers) during the training phase. TapNet \cite{Zhang_Gao_Lin_Lu_2020} and DVSL \cite{10.1145/3340531.3412099} are explicit prototypes contrast because explicit prototypes are introduced. TapNet introduces a learnable prototype for each predefined class and classifies the input time series sample according to the distance between the sample and each class prototype. DVSL defines virtual sequences, which have the same function as prototypes, i.e., minimize the distance between samples and virtual sequences, but maximize the distance between virtual sequences. MHCCL \cite{meng2022mhccl} proposes a hierarchical clustering based on the upward masking strategy and a contrastive pairs selection strategy based on the downward masking strategy. In the upward mask strategy, MHCCL believes that outliers greatly impact prototypes, so these outliers should be removed when updating prototypes. The downward masking strategy, in turn, uses the clustering results to select positive and negative samples, i.e., samples belonging to the same prototype are regarded as true positive samples, and samples belonging to different prototypes are regarded as true negative samples.

\vspace{-5pt}

\subsection{Expert knowledge contrast}
Expert knowledge contrast is a relatively new representation learning framework. Generally speaking, this modeling framework incorporates expert prior knowledge or information into deep neural networks to guide model training \cite{WU2022364,ikdml2022}. In the contrastive learning framework, prior knowledge can help the model choose the correct positive and negative samples during training. An example of expert knowledge contrast is shown in Fig. \ref{fig:contrastshow}(e). 

Here we sort out three typical works of expert knowledge contrast for time series data. Shi et al. \cite{9533426} used the DTW distance of time series samples as prior information and believed that two samples with a small DTW distance have a higher similarity. Specifically, given the anchor $x^{ref}$ and the other two samples $x_i$ and $x_j$, the DTW distance between $x^{ref}$ and the other two samples is calculated first, then the sample with a small distance from $x^{ref}$ is considered as the positive sample of $x^{ref}$. This selection process is defined as
\begin{equation}
	\text{label} =\left\{\begin{array}{ll}
		1, & \mathrm{DTW}\left(\boldsymbol{x}^{ref}, \boldsymbol{x}_{i}\right) \geq \operatorname{DTW}\left(\boldsymbol{x}^{ref}, \boldsymbol{x}_{j}\right) \\
		0, & \text { otherwise }
	\end{array}. \right.
	\label{eq:shidtwlabel}
\end{equation}
Based on pair-loss, ExpCLR\cite{pmlr-v162-nonnenmacher22a} introduces expert features of time series data to obtain more informative representations. Given two input samples $x_i$ and $x_j$ and corresponding representations $f_i$ and $f_j$, ExpCLR defines the normalized distance between two samples:
\begin{equation}
	s_{ij} = 1 - \frac{\left \| f_i-f_j \right \|_2}{\mathtt{max}\left \|f_k-f_l\right \|}_2,
\end{equation}
where $f_k$ and $f_l$ are the two representation vectors with the largest distance among all samples. Compared with the original pair-loss, the distance between samples $x_i$ and $x_j$ is changed from a discrete value (0 and 1) to a continuous value $s_{ij}$, which enables the model to learn more accurately about the relationship between samples, thus thereby enhancing the representation ability of the model. In addition to the above two works, SleepPriorCL\cite{sleeppriorcl} was proposed to alleviate the sampling bias problem faced by (\ref{eq:simclr}). Like ExpCLR, SleepPriorCL also introduces prior features to ensure the model can identify correct positive and negative samples.

Actually, introducing more prior knowledge in contrastive-based SSL can help the model extract better representations. The trend of this family of methods can be summarized from two perspectives: (i) Addressing sampling bias. Sampling bias is caused by inappropriate selection of positive and negative samples, so introducing prior knowledge useful for selecting positive and negative samples can deal with this problem, such as a clustering-based negative sample detection algorithm \cite{DBLP:journals/corr/abs-2106-03719} and sample identification strategy based on real labels \cite{NEURIPS2020_d89a66c7,pmlr-v139-yeche21a}. (ii) Addressing representation bias. Representation bias means that the extracted representations cannot be guaranteed to be strongly related to the downstream task. The essential reason is that there may be a big difference between the goals of the pretext task and the downstream task. An interesting trend is to fuse semi-supervised learning and contrastive-based SSL to guide the training of the encoder through a small amount of labeled data \cite{ZHANG20233197,9732218}.


\section{Adversarial-based Methods}
\label{sec:adversarial}
Adversarial-based self-supervised representation learning methods utilize generative adversarial networks (GANs) to construct pretext tasks. GAN contains a generator $\mathcal{G}$ and a discriminator $\mathcal{D}$. The generator $\mathcal{G}$ is responsible for generating synthetic data similar to real data, while the discriminator $\mathcal{D}$ is responsible for determining whether the generated data is real data or synthetic data. Therefore, the goal of the generator is to maximize the decision failure rate of the discriminator, and the goal of the discriminator is to minimize its failure rate \cite{NIPS2014_5ca3e9b1,10.1145/3559540}. The generator $\mathcal{G}$ and the discriminator $\mathcal{D}$ are a mutual game relationship, so the learning objective is:
\begin{equation}
	\mathcal{L} = \mathbb{E}_{x \sim \mathcal{P}_{\text{data}}(x)}[\log \mathcal{D}(x)]+\mathbb{E}_{z \sim \mathcal{P}_{\mathbf{z}}(z)}[\log (1-\mathcal{D}(\mathcal{G}(\mathbf{z})))].
	\label{eq:ganloss}
\end{equation}

According to the final task, the existing adversarial-based representation learning methods can be divided into time series generation and imputation, and auxiliary representation enhancement. The illustration of the adversarial-based SSL for time series is shown in Fig. \ref{fig:advshow}. In Appendix~\ref{app:tsgi} - \ref{app:are}, the main advantages and disadvantages of two adversarial-based submethods are summarized. Furthermore, the main differences in characteristics and limitations between the adversarial-based methods and the previous two methods (generative-based and contrastive-based) are shown in Appendix~\ref{app:charlimi}.

        
        
		

\begin{figure}[!t]
	\centering
	\subfloat[Time series generation]{\includegraphics[width=75mm]{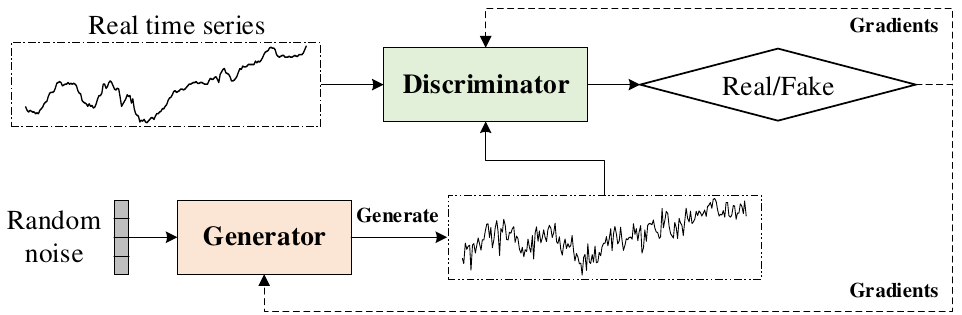}} \\
	\subfloat[Time series imputation]{\includegraphics[width=75mm]{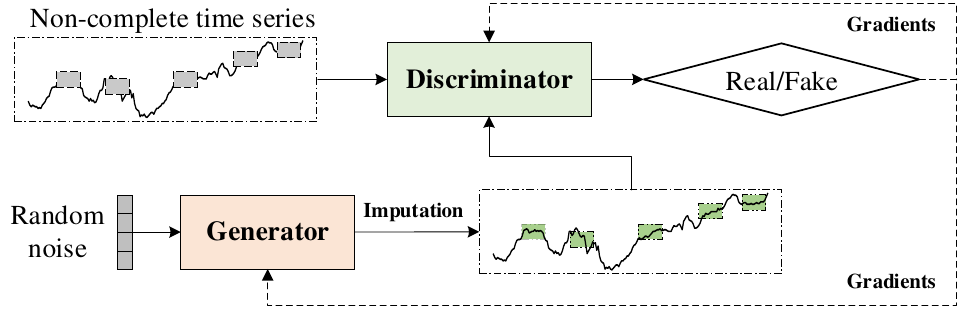}} \\
	\subfloat[Auxiliary representation enhancement]{\includegraphics[width=75mm]{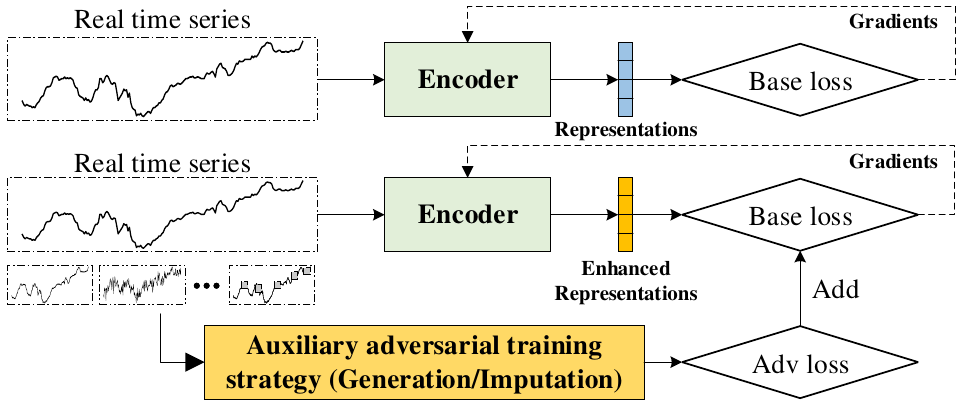}} \\
	\caption{Three categories of adversarial-based SSL for time series data.}
	\label{fig:advshow}
    \vspace{-10pt}
\end{figure}

    


\subsection{Time series generation and imputation}

The generator in the GAN can generate synthetic data close to the real data, so adversarial representation learning has a wide range of applications in the data generation field \cite{10.1145/3439723}, especially in image generation \cite{8100115,8237506,brock2018large,8953766}. In recent years, many scholars have also explored the potential of generative representation learning in time series generation and imputation, such as C-RNN-GAN \cite{DBLP:journals/corr/Mogren16}, TimeGAN \cite{10.5555/3454287.3454781}, TTS-GAN \cite{lixiaominttsgan}, and E$^2$GAN \cite{ijcai2019p429}. It should be emphasized that although Brophy et al. \cite{10.1145/3559540} have reviewed the GAN-based time series generation methods in the latest survey, it differs from the proposed taxonomy. We sort out the two aspects of complete time series generation and missing value imputation, while Brophy et al. sorted out from the perspective of discrete and continuous time series modeling.

Complete time series generation refers to generating a new time series that does not exist in the existing data set. The new sample can be a univariate or multivariate time series. C-RNN-GAN \cite{DBLP:journals/corr/Mogren16} is an early method of generating time series samples using GAN. The generator is an RNN, and the discriminator is a bidirectional RNN. RNN-based structures can capture the dynamic dependencies in multiple time steps but ignore the static features of the data. TimeGAN \cite{10.5555/3454287.3454781} is an improved time series generation framework that combines the basic GAN with the autoregressive model, allowing the preservation of temporal dynamic characteristics of the series. TimeGAN also emphasizes that static features and temporal characteristics are crucial to the generation task. 

Some recently proposed methods consider more complex time series generative tasks\cite{seyfi2022generating,jeha2022psagan,lixiaominttsgan,jeon2022gtgan}. For example, COSCI-GAN \cite{seyfi2022generating} is a time series generation framework that considers the correlation between each dimension of the multivariate time series. It includes Channel GANs and Central Discriminator. Channel GANs are responsible for generating data in each dimension independently, while Central Discriminator is responsible for determining whether the correlation between different dimensions of the generated series is the same as the raw series. PSA-GAN \cite{jeha2022psagan} is a framework for long-time series generation and introduces a self-attention mechanism. It further presents Context-FID, a new metric for evaluating the quality of generated series. Li et al. \cite{lixiaominttsgan} explored the generation of time series data with irregular spatiotemporal relationships and proposed TTS-GAN, which uses a Transformer instead of an RNN to build the discriminator and the generator and treats the time series data as image data of height one. 


Different from generating a new time series, the task of time series imputation refers to that given a non-complete time series sample (for example, the data of some time steps is missing), and the missing values need to be filled based on the contextual information. Luo et al. \cite{NEURIPS2018_96b9bff0} treat the problem of missing value imputation as a data generation task and then use GAN to learn the distribution of the training data set. In order to better capture the dynamic characteristics of the series, the GRUI module was proposed. The GRUI uses the time-lag matrix to record the time-lag information between effective values of incomplete time series data, which follow the unknown non-uniform distribution and are very helpful for analyzing the dynamic characteristics of the series. The GRUI module was also further used in E$^2$GAN \cite{ijcai2019p429}. SSGAN \cite{Miao_Wu_Wang_Gao_Mao_Yin_2021} is a semi-supervised framework for time series data imputation, which includes a generative network, a discriminative network, and a classification network. Unlike previous frameworks, SSGAN's classification network makes full use of label information, which helps the model achieve more accurate imputations.


\subsection{Auxiliary representation enhancement}

In addition to generation and interpolation tasks, an adversarial-based representation learning strategy can be added to existing learning frameworks as additional auxiliary learning modules, which we call adversarial-based auxiliary representation enhancement. The auxiliary representation enhancement aims to promote the model to learn more informative representations for downstream tasks by adding adversarial-based learning strategies. It can be defined as:
\begin{equation}
	\mathcal{L} = \mathcal{L}_{base} + \mathcal{L}_{adv},
	\label{eq:gan3}
\end{equation}
where $\mathcal{L}_{base}$ is the basic learning objective and $\mathcal{L}_{adv}$ is the additional adversarial-based learning objective. It should be noted that when $\mathcal{L}_{adv}$ is not available, the model can still extract representations from the data, so $\mathcal{L}_{adv}$ is regarded as an auxiliary learning objective.

USAD \cite{10.1145/3394486.3403392} is a time series anomaly detection framework that includes two BAE models, and two BAE are defined as $AE_1$ and $AE_2$, respectively. The core idea behind USAD is to amplify the reconstruction error by adversarial training between two BAEs. In USAD, $AE_1$ is regarded as the generator, and $AE_2$ is regarded as the discriminator. The auxiliary goal is to use $AE_2$ to distinguish real data from reconstructed data from $AE_1$, and train $AE_1$ to deceive $AE_2$, the whole process can be expressed as:
\begin{equation}
	\mathcal{L}_{adv} = \min_{AE_1}\max_{AE_2} \left \| {W - AE_2(AE_1(W))} \right \|_2,
	\label{eq:usad}
\end{equation}
where $W$ is the real input series. Similar to USAD, AnomalyTrans \cite{xu2022anomaly} also uses an adversarial strategy to amplify the anomaly score of anomalies. But unlike (\ref{eq:usad}), which uses reconstruction error, AnomalyTrans defines prior-association and series-association and then uses the Kulback-Leibler divergence to measure the error of the two associations. 

DUBCNs \cite{Zhu_Song_Chen_Lumezanu_Cheng_Zong_Ni_Mizoguchi_Yang_Chen_2020} and CRLI \cite{Ma_Chen_Li_Cottrell_2021} are used for series retrieval and clustering tasks, respectively. Both methods adopt RNN-based BAE as the model, and the clustering-based loss and adversarial-based loss are added to the basic reconstruction loss, i.e., 
\begin{equation}
	\mathcal{L} = \mathcal{L}_{mse} + \lambda_1 \mathcal{L}_{cluster} + \lambda_2 \mathcal{L}_{adv}.
	\label{eq:DUBCNs}
\end{equation}
where $\lambda_1$ and $\lambda_2$ are the weight coefficients of the auxiliary objective.

The adversarial-based strategy is also effective in other time series modeling tasks. For example, introducing adversarial training in time series forecasting can improve the accuracy and capture long-term repeated patterns, such as AST \cite{10.5555/3495724.3497159} and ACT \cite{9892791}. BeatGAN \cite{ijcai2019p616} introduces adversarial representation learning in the abnormal beat detection task of ECG data and provides an interpretable detection framework. In modeling behavior data, Activity2vec \cite{Aggarwal_Joty_Fernandez-Luque_Srivastava_2019} uses adversarial-based training to model target invariance and enhance the representation ability of the model in different behavior stages.


\section{Applications and Datasets}
\label{sec:applications}

\begin{table} 
    \caption{Summary of time series applications and widely used datasets. The UCR and UEA datasets have multiple sub-datasets respectively, so their size and dimension are not fixed values. The size and dimension of each sub-dataset are represented by $\mathcal{M}$ and $\mathcal{D}$, respectively. AnRa represents \textbf{\underline{An}}omaly \underline{Ra}tio. SaIn represents \textbf{\underline{Sa}}mpling \underline{In}terval. Datasets for classification tasks and clustering tasks are listed together because the goals of these two tasks are similar. AD represents anomaly detection. F represents forecasting. C$\&$C represents classification and clustering.}  
    \label{tab:appdata}
    \centering
    \setlength{\tabcolsep}{1.2mm}{
    \begin{tabular}{c|cccc} \toprule\toprule	
	App & Dataset & Size & Dim  & Comment\\ \midrule
	
	\multirow{6}{*}{AD} 
 	& PSM$^{\href{https://github.com/eBay/RANSynCoders}{*}}$ \cite{10.1145/3447548.3467174}
        & 132,481 / 87,841 & 26
        & AnRa: 27.80\% \\
        
        
	& SMD$^{\href{https://github.com/NetManAIOps/OmniAnomaly}{*}}$ \cite{10.1145/3292500.3330672}
        & 708,405 / 708,405 & 38 
        & AnRa: 4.16\% \\
        
        & MSL$^{\href{https://github.com/khundman/telemanom}{*}}$ \cite{10.1145/3219819.3219845}
        & 58,317 / 73,729 & 55 
        & AnRa: 10.72\% \\ 
        
	& SMAP$^{\href{https://github.com/khundman/telemanom}{*}}$ \cite{10.1145/3219819.3219845}
        & 135,183 / 427,617 & 25
        & AnRa: 13.13\% \\
        
	& SWaT$^{\href{https://itrust.sutd.edu.sg/itrust-labs_datasets/dataset_info/}{*}}$  \cite{goh2017dataset}
        & 475,200 / 449,919 & 51
        & AnRa: 12.98\% \\ 
        
	& WADI$^{\href{https://itrust.sutd.edu.sg/itrust-labs_datasets/dataset_info/}{*}}$ \cite{10.1145/3055366.3055375}
        & 1,048,571 / 172,801 & 103
        & AnRa: 5.99\% \\ 
        
	\midrule
	
	\multirow{9}{*}{F} 
	& ETTh$^{\href{https://github.com/zhouhaoyi/ETDataset}{*}}$ \cite{Zhou_Zhang_Peng_Zhang_Li_Xiong_Zhang_2021} 
        & 17,420 & 7 
        & SaIn: 1h \\

        & ETTm$^{\href{https://github.com/zhouhaoyi/ETDataset}{*}}$ \cite{Zhou_Zhang_Peng_Zhang_Li_Xiong_Zhang_2021} 
        & 69,680 & 7 
        & SaIn: 15min \\

	& Wind$^{\href{https://www.kaggle.com/datasets/sohier/30-years-of-european-wind-generation}{*}}$ \cite{ecsp_wind} 
        & 10,957 & 28
        & SaIn: 1day \\
 
	& Electricity$^{\href{https://archive.ics.uci.edu/ml/datasets/ElectricityLoadDiagrams20112014}{*}}$ \cite{uci_electr}
        & 26,304 & 321
        & SaIn: 1hour\\
 

        & ILI$^{\href{https://gis.cdc.gov/grasp/fluview/fluportaldashboard.html}{*}}$ \cite{ilidata} 
        & 966 & 7
        & SaIn: 1weak\\

        & Weather$^{\href{https://www.bgc-jena.mpg.de/wetter/}{*}}$ \cite{weatherdata} 
        & 52,696 & 21
        & SaIn: 10min\\

        & Traffic$^{\href{http://pems.dot.ca.gov/}{*}}$ \cite{traffic} 
        & 17,544 & 862 
        & SaIn: 1hour\\

        & Exchange$^{\href{https://github.com/laiguokun/LSTNet}{*}}$ \cite{10.1145/3209978.3210006}
        & 7,588 & 8 
        & SaIn: 1day\\

	& Solar$^{\href{https://www.nrel.gov/grid/solar-power-data.html}{*}}$ \cite{nrel_solar} 
        & 52,560 & 137
        & SaIn: 10min\\

        \midrule 
	
	\multirow{3}{*}{C$\&$C} 

        & HAR$^{\href{https://archive.ics.uci.edu/ml/datasets/human+activity+recognition+using+smartphones}{*}}$ \cite{Anguita2013APD}& 17,3056 / 173,056 & 9
        & Classes: 6\\
        
	& UCR 128$^{\href{https://www.cs.ucr.edu/~eamonn/time_series_data/}{*}}$ \cite{8894743} 
        & 128 * $\mathcal{M}$ & 1
        & N/A \\
        
	& UEA 30$^{\href{https://www.timeseriesclassification.com/}{*}}$ \cite{bagnall2018uea} 
        & 30 * $\mathcal{M}$ & $\mathcal{D}$ 
        & N/A \\

    \bottomrule\bottomrule
    \end{tabular}
}
\vspace{-10pt}
\end{table}

SSL has many applications across different time series tasks. This section summarizes the most widely used datasets and representative references according to the application area, including anomaly detection, forecasting, classification, and clustering. As shown in Table \ref{tab:appdata}, we provide useful information, including dataset name, dimension, size, source, and useful comments. For each task, we summarize from the following aspects: task description, related methods, evaluation metrics, examples, and task flow. Due to space limitations, relevant descriptions of evaluation metrics, examples, and task flow can be found in Appendix~\ref{app:app}. In addition, we provide performance comparison results of different methods on the same dataset and further summarize the correlation between methods and tasks, the details can also be found in Appendix~\ref{app:qca}.

\subsection{Anomaly detection}

\begin{itemize}[leftmargin=*]
\item \textbf{Task description.} The anomaly detection problem for time series is usually formulated as identifying outlier time points or unexpected time sequences relative to some norm or usual signal.

\item \textbf{Related methods.} Most time series anomaly detection methods are constructed under an unsupervised learning framework because obtaining labels for anomalous data is challenging. Autoregressive-based forecasting and autoencoder-based reconstruction are the most commonly used modeling strategies. To be concrete, THOC \cite{NEURIPS2020_97e401a0} and GDN \cite{Deng_Hooi_2021} employ autoregressive-based forecasting SSL framework, which assumes that anomalous sequences or time points are not predictable. RANSynCoders \cite{10.1145/3447548.3467174}, USAD \cite{10.1145/3394486.3403392}, AnomalyTrans \cite{xu2022anomaly}, and DAEMON \cite{10.1145/3539597.3570371} employ autoencoder-based reconstruction SSL framework. Furthermore, VGCRN \cite{pmlr-v162-chen22x} and FuSAGNet \cite{10.1145/3534678.3539117} combine two frameworks to achieve more robust and accurate results. It is beneficial to introduce an adversarial-based SSL, which can further amplify the difference between normal and anomalous data, such as USAD \cite{10.1145/3394486.3403392} and DAEMON \cite{10.1145/3539597.3570371}.

\end{itemize}


\subsection{Forecasting}

\begin{itemize}[leftmargin=*]
\item \textbf{Task description.} Time series forecasting is the process of analyzing time series data using statistics and modeling to make predictions of future windows or time points. 

\item \textbf{Related methods.} The pretext task based on autoregressive-based forecasting is essentially a time series forecasting task. Therefore, various models based on forecasting tasks are proposed, such as Pyraformer \cite{liu2022pyraformer}, FilM \cite{zhou2022film}, Quatformer \cite{ 10.1145/3534678.3539234}, Informer \cite{Zhou_Zhang_Peng_Zhang_Li_Xiong_Zhang_2021}, Triformer \cite{ijcai2022p277}, Scaleformer\cite{shabani2023scaleformer}, Crossformer \cite{zhang2023crossformer}, and Timesnet \cite{wu2023timesnet}. Moreover, we found that decomposing the series (seasonality and trend) and then learning and forecasting on the decomposed components will help improve the final forecasting accuracy, such as MICN \cite{wang2023micn} and CoST \cite{woo2022cost}. Besides, introducing an adversarial SSL is viable when missing values are in the series. For example, LGnet \cite{tangsuncharu2020} introduces adversarial training to enhance the modeling of global temporal distribution, which mitigates the impact of missing values on forecasting accuracy.


\vspace{-5pt}
\end{itemize}

\subsection{Classification and clustering} 

\begin{itemize}[leftmargin=*]
\item \textbf{Task description.} The goal of classification and clustering tasks is similar, i.e., to identify the real category to which a certain time series sample belongs.

\item \textbf{Related methods.} Contrastive-based SSL methods are the most suitable choice for these two tasks since the core of contrastive learning is identifying positive and negative samples. Specifically, TS-TCC \cite{ijcai2021-324} introduces temporal contrast and contextual contrast in order to obtain more robust representations. TS2Vec \cite{Yue_Wang_Duan_Yang_Huang_Tong_Xu_2022} and MHCCL \cite{meng2022mhccl} perform a hierarchical contrastive learning strategy over augmented views, which enables robust representations. Similar to anomaly detection and prediction tasks, an adversarial-based SSL strategy can also be introduced into classification and clustering tasks. DTCR \cite{NEURIPS2019_1359aa93} propose a fake-sample generation strategy to assist the encoder in obtaining more expressive representations.



\end{itemize}

\vspace{-5pt}
\section{Discussion and Future Directions}
\label{sec:future}
In this section, we point out some critical problems in current studies and outline several research directions worthy of further investigation. 

\subsection{Selection and combination of data augmentation}

Data augmentation is one of the effective methods to generate augmented views in SSCL \cite{10.1007/s00521-020-04748-3,ncaaug2023}. The widely used methods for time series data include jitter, scaling, rotation, permutation, and warping \cite{POPPELBAUM2022108397,ijcai2021p631,9539005,Iwana_2021,gao2020robusttad}. In SimCLR \cite{10.5555/3524938.3525087}, nine different augmentation methods for image data were discussed. The experiments show that ``no single transformation suffices to learn good representations" and ``the composition of random cropping and random color distortion is the most effective augmentation method". This naturally raises the question of which one or composition of data augmentation methods is optimal for time series. Recently, Um et al. \cite{TerryUm_ICMI2017} show that the combination of three basic augmentation methods (permutation, rotation, and time warping) is better than that of a single method and achieves the best performance in time series classification task. Iwana et al. \cite{Iwana_2021} evaluate twelve time series data augmentation methods on 128 time series classification datasets with six different types of neural networks.  Different evaluation frameworks give different recommendations and results. Therefore, an interesting direction is to construct a reasonable evaluation framework for time series data augmentation methods, then further select the optimal method or combination strategy.

\subsection{Inductive bias for time series SSL}
Existing SSL methods often pursue an entirely data-driven modeling approach. However, introducing reasonable inductive bias or prior is helpful for many deep neural networks-based modeling tasks \cite{WU2022364,DENG2020101656,chen2022integration}. On the one hand, although a purely data-driven model can be easily extended to various tasks, it requires much data to train it. On the other hand, time series data usually has some available characteristics, such as seasonal, periodic, trend, and frequency domain biases~\cite{wen2019robuststl,wen2021robustperiod,zhou2022fedformer}. Thus one future direction is to consider more effective ways to induce inductive biases into time series SSL based on the understanding of time series data and characteristics of specific tasks.

\subsection{SSL for irregular and sparse time series }
Irregular and sparse time series also widely exist in various scenarios. This data is measured at irregular time intervals, and not all the variables are available for each sample \cite{DBLP:journals/corr/abs-2012-00168}. The straightforward approach to deal with irregular and sparse time series data is to use interpolation algorithms to estimate missing values \cite{NEURIPS2018_96b9bff0,Miao_Wu_Wang_Gao_Mao_Yin_2021,Wu_Ni_Cheng_Zong_Song_Chen_Liu_Zhang_Chen_Davidson_2021}. However, interpolation-based models add undesirable noise and extra overhead to the model which usually worsens as the time series become increasingly sparse \cite{10.1145/3516367}. Moreover, irregular and sparse time series data is often expensive to obtain sufficient labeled data, which motivates us to build time series analysis models based on SSL in various tasks. Therefore, building SSL models directly on irregular and sparse time series data without interpolation is a valuable direction.

\subsection{Pretraining and large models for time series}
Nowadays, many large language models have shown their powerful perception and learning capability for many different tasks. A similar phenomenon also appears in computational vision~\cite{kirillov2023segany}. It is naturally an interesting question, how about the time series analysis field? As far as we know, there is limited work on pretraining models in large-scale time series. Exploring the potentiality of pretraining and large time series models is a promising direction.

\subsection{Adversarial attacks and robust analysis on time series}
With the widespread use of deep neural networks in time series forecasting, classification, and anomaly detection, the vulnerability and robustness of deep models under adversarial have become a significant concern \cite{ipnn2013,liu2023robust,WU2022794,9063523}. In the field of time series forecasting, Liu et al. \cite{liu2023robust} study the indirect and sparse adversarial attacks on multivariate probabilistic forecasting models for time series forecasting and propose two defense mechanisms: randomized smoothing and mini-max defense. Wu et al. \cite{WU2022794} propose an attack strategy for generating an adversarial time series by adding malicious perturbations to the original time series to deteriorate the performance of time series prediction models.  Zhuo et al. \cite{9852307} summarize and compare various recent and typical adversarial attack and defense methods for fault classifiers in data-driven fault detection and classification systems, including white-box attack (FGSM\cite{goodfellow2015explaining}, IGSM\cite{lee2017making}, C$\&$W attack\cite{7958570}, DeepFool\cite{7780651}) and gray-box and black-box attack (UAP \cite{8099500}, SPSA \cite{pmlr-v80-uesato18a}, Random noise). The research on adversarial attacks and defenses against time series data is a worthwhile direction, but there is much less literature on this topic. Existing studies mainly involve forecasting and classification tasks. However, the impact of adversarial examples on time series self-supervised pre-training tasks is still unknown.


\subsection{Benchmark evaluation for time series SSL}
SSL has many applications in time series classification, forecasting, clustering, and anomaly detection. However, most current research seeks to achieve the best performance on specific tasks and needs more discussion and evaluation of the self-supervised technique. One interesting direction is to pay more attention to SSL, analyze its properties in time series modeling tasks, and give reliable benchmark evaluation.

\subsection{Time series SSL in collaborative systems}
Distributed systems have been widely deployed in many scenarios, including intelligent control systems, wireless sensor networks, network file systems, etc. On the one hand, an appropriate collaborative learning strategy is fundamental in these systems, as users can train their own local models without sharing their private local data and circumventing the relevant privacy policy \cite{li2023mocosfl}. On the other hand, time series data is also widely distributed in various places in the system, and obtaining sufficient labeled data is also difficult, so time series SSL has great deployment potential. In recent years, federated learning has been the most popular collaborative learning framework and has been used successfully in various applications. Combining time series self-supervised learning and federated learning is a valuable research direction that can provide additional modeling tools for modern distributed systems.

\section{Conclusion}
\label{sec:con}
This article concentrates on time series SSL methods and provides a new taxonomy. We categorize the existing methods into three broad categories according to their learning paradigms: generative-based, contrastive-based, and adversarial-based. Moreover, we sort out all methods into ten detailed subcategories: autoregressive-based forecasting, autoencoder-based reconstruction, diffusion-based generation, sampling contrast, prediction contrast, augmentation contrast, prototypes contrast, expert knowledge contrast, generation and imputation, and auxiliary representation enhancement. We also provide useful information about applications and widely used time series datasets. Finally, multiple future directions are summarized. We believe this review fills the gap in time series SSL and ignites further research interests in SSL for time series data.



\ifCLASSOPTIONcaptionsoff
  \newpage
\fi

\bibliographystyle{IEEEtran}
\bibliography{refs}

\newpage
\appendices

\section{Learning paradigms of SSL}
\label{app:lp}
\begin{figure*}[!t]
	\centering
	\includegraphics[width=185mm]{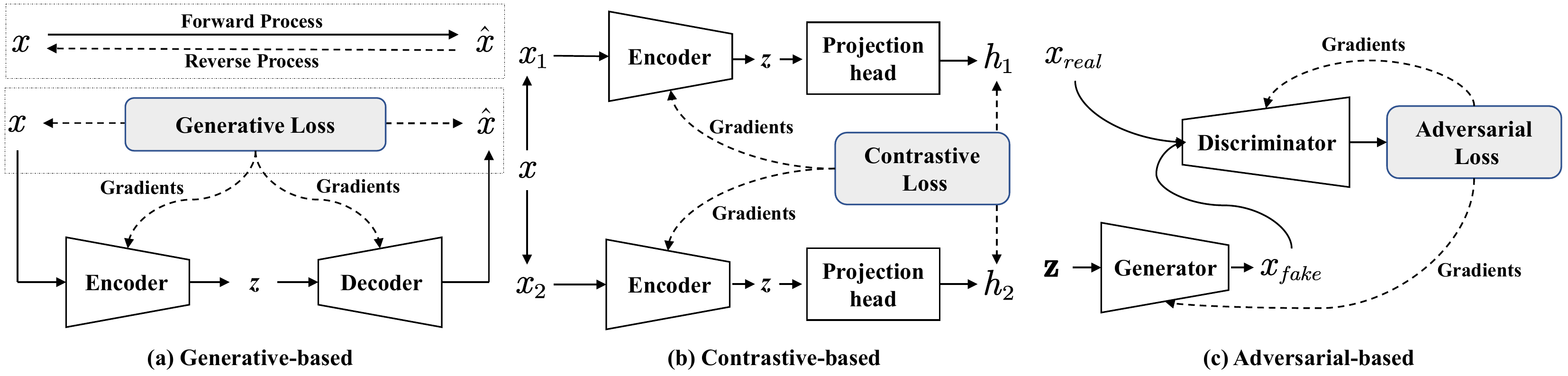}
	\caption{Learning paradigms of SSL.}
	\label{fig:sslframework}
\end{figure*}

The model architectures of generative-based, contrastive-based, and adversarial-based methods are shown in Fig. \ref{fig:sslframework} and can be summarized as: 

\begin{itemize}[leftmargin=*]
    \item The generative-based approach first uses an encoder to map the input $\mathbf{x}$ to the representation $\mathbf{z}$, and then a decoder to reconstruct $\mathbf{x}$ from $\mathbf{z}$. The training objective is to minimize the reconstruction error between the input $\mathbf{x}$ and the reconstructed input $\hat{\mathbf{x}}$. 
    
    
    \item The contrastive-based approach is one of the most widely used SSL strategies, which constructs positive and negative samples through data augmentation or context sampling. The model is then trained by maximizing the Mutual Information (MI) between the two positive samples. Contrastive-based methods usually use a contrastive similarity metric, such as InfoNCE loss \cite{DBLP:journals/corr/abs-1807-03748}. 
    
    \item The adversarial-based approach usually consists of a generator and a discriminator. The generator generates fake samples, and the discriminator is used to distinguish them from real samples. 
    
\end{itemize}

\section{Further Description and Summary of Four Pipelines of SSL}

\subsection{Summary regarding positive and negative samples}
\label{app:posneg}

\begin{itemize}[leftmargin=*]

    \item Multisensory signals: One direct approach to construct positive and negative samples is to record the information with multiple sensors. These sensors can be of the same modality or of different modalities.
    
    \item Data augmentation: The most simple yet effective method is to generate positive and negative samples through hand-crafted transformation rules, often called data augmentation. The choice of rules depends on the data modality. For example, the augmentation methods of image data include crop, color distort, resize, and rotate \cite{10.5555/3524938.3525087}, while for time series, the augmentation methods involve noise injection, window slicing, and window wrapping \cite{ijcai2021p631}. 

    \item Local-global consistency: Explicitly constraining the representation of the local view to be similar to the representation of the global view, while being different from the representation of other instances is also an effective approach for generating positive and negative samples.
    
    \item Temporal consistency: Using temporal dependence to construct positive and negative samples is a common method in sequential data, such as time series and video data. Generally, the representation of continuous or close views in a sequence is considered as a positive pair while discontinuous and far away pairs in the same sequence or different sequences are considered negative pairs.
    
\end{itemize}

\subsection{Further classification of pretexts}
\label{app:pretext}

The main differences of three types of pretext tasks are as follows:

\begin{itemize}[leftmargin=*]
    \item Context prediction: This family of methods uses the inherent context relationship of the provided instances to construct pretext tasks. This relationship can involve spatial structure, temporal structure, and local and global consistency. The model is trained to predict the contextual structure of the instance itself.
    
    \item Instance discrimination: In this family of methods, the pretext tasks focus on the similarities between different views or modalities of the same instance. Each instance is treated as its own class, or assigning pseudo-labels to training instances through natural clustering and dividing the training instances into a number of groups with high-intragroup and low-intergroup similarity. The model is trained to discriminate between different instances or groups.
    

    \item Instance generation: This family of methods constructs pretext tasks by reconstructing perturbed instances or generating instances through models based on generative adversarial networks (GANs). The core is to regenerate a new instance based on a model or rule, and use the predefined relationship between the original instance of the newly generated instance to construct pseudo-supervision signals.
\end{itemize}

\begin{table}[!t]
    \caption{Classification of commonly used pretext tasks.}
    \label{table-pretext}
    \centering
    \begin{tabular}{c|c} \toprule\toprule
    Category & Pretext tasks  \\ \midrule
    \multirow{4}{*}{Context prediction} 
    & Spatial transformation \cite{9770283,gui2023survey,technologies9010002}\\ 
    & Temporal prediction \cite{9086055,technologies9010002}\\
    & Masked prediction \cite{9770283} \\
    & Cross-modal-view prediction \cite{technologies9010002}\\

    \midrule
    
    \multirow{5}{*}{Instance discrimination} 
    & Clustering based \cite{9770283,9086055}  \\
    & Regularization based \cite{9770283} \\
    & Negative example SSCL based \cite{gui2023survey}\\
    & Self-distillation SSCL based \cite{gui2023survey}\\
    & Feature decorrelation SSCL based \cite{gui2023survey}\\

    \midrule
    
    \multirow{2}{*}{Instance generation} 
    & Perturbation and reconstruction \cite{9770382} \\
    & GAN-based instance generation \cite{9086055} \\
    \bottomrule \bottomrule
\end{tabular}

\end{table}

\subsection{Characteristics of four SSL model architectures}
\label{app:modelarchitecture}

\begin{itemize}[leftmargin=*]
    \item End-to-End: This architecture utilizes two encoders of the same structure to encode different views of the instance, and then updates the two encoders simultaneously based on the contrastive loss. This architecture prefers large batch sizes to accumulate a greater number of negative samples.
    
    \item Memory bank: Maintaining a large training batch places high demands on computing resources. Memory bank is a possible solution, which constructs a dictionary that stores and updates the embeddings of samples with the most recent ones at regular intervals. However, the consistency of the representations cannot be guaranteed.
    
    \item Momentum encoder: To address the issues with End-to-End and Memory bank architectures, using a momentum encoder which acts as a dynamic dictionary lookup for encodings of negative samples during training is a possible architecture.

    \item Clustering: Another way to reduce the burden of a large training batch is to introduce a clustering mechanism. This architecture not only makes a pair of samples close to each other but also makes sure that all other features that are similar to each other form clusters together.
    
\end{itemize}

\subsection{Objectives of four SSL losses}
\label{app:sslloss}
\begin{itemize}[leftmargin=*]
    \item Scoring functions: The scoring function measures compatibility between two vectors either in terms of similarity or distance. Depending on the specific loss function, for positive pairs either the similarity score is maximized or the distance metric is minimized.

    \item Energy-based margin functions: Energy-based models are a general class of models that associate an energy (distance score) with each configuration of the variables to be modeled (pairs of query and keys vectors). This family of loss functions involves associating a low energy (small distance) to desired configurations of the variable (positive pairs) and high energy to undesired configurations of variables (negative pairs).

    \item Probabilistic NCE-based functions: This family of functions is a non-parametric version for the softmax function that correctly identifies the positive for a given query and contains all negative keys with one one positive key.

    \item Mutual information based functions: Motivated by mutual information, this family maximizes the mutual information between representations of different views of the same instance.
    
\end{itemize}

\label{app:sumgen3}
\begin{table*}[!t]
    \caption{Summary of advantages and disadvantages of three generative-based submethods.}
    \label{table-gen-summary}
    \centering
    \begin{tabular}{c|cc} \toprule\toprule
    Method & Advantages & Disadvantages\\ 
    \midrule
    \multirow{2}{*}{Autoregressive-based forecasting} 
    & Preserve inherent temporal dependencies & Inaccurate forecasting of extreme points \\ 
    & Capture dynamic features and patterns & Error accumulation in long-term forecasting \\ 
    \midrule
    
    \multirow{2}{*}{Autoencoder-based reconstruction}
    & Flexible structure (BAE, DAE, MAE, VAE, etc)  & Difficulty in extracting temporal dependencies\\
    & Extract robust and low-dimensional representations & Struggle with complex time series data\\

    \midrule
    \multirow{2}{*}{Diffusion-based generation}
    & Concise and flexible workflow & Difficulty in obtaining semantic information \\
    & Potential for integration with other models & Inefficient sampling and likelihood estimation\\
    \bottomrule \bottomrule
\end{tabular}

\vspace{-10pt}
\end{table*}

\section{Advantages and Disadvantages of Three Generative-based Methods}
\label{app:adgen}

\subsection{Autoregressive-based forecasting}
\label{app:arf}
 Using autoregressive-based forecasting tasks as a pretext task for time series SSL has the following advantages. (i) Autoregressive-based forecasting tasks leverage the inherent structure of raw time series without the need for manually annotated labels. It is easy to construct an abundant training data set through slicing operations. (ii) The inherent structure of time series data allows the autoregressive forecasting task to directly build upon the temporal dependencies in the data, allowing the representations to capture the dynamic characteristics and patterns of the time series. Their main disadvantages are as follows. (i) The objective of autoregressive-based forecasting tasks is to forecast future values of a time series, which can be constrained by the predictability of the target and the availability of data. If the sequence exhibits high uncertainty or sudden events, the autoregressive prediction model may struggle to accurately forecast future values, thereby affecting the effectiveness of SSL. (ii) Forecasting errors in autoregressive models may accumulate over time, especially in long-term prediction tasks. This can lead to the extracted representations exhibiting bias over time.

\subsection{Autoencoder-based reconstruction}
\label{app:aer}
 The advantages are as follows. (i) Autoencoder-based reconstruction is the easiest-to-use pretext task in SSL, especially MAE models. It can be used for tasks such as feature learning and dimensionality reduction.  (ii) Autoencoders are often trained by adding noise or applying random transformations. This enables the model to better handle noise and uncertainties in the time series data during the reconstruction process, thereby enhancing representation robustness. Their disadvantages are as follows. (i) Autoencoder-based reconstruction tasks face challenges in handling long-term dependencies in time series data. A single reconstruction constraint cannot guarantee that the extracted representations capture the temporal dependencies in time series data. (ii) Time series data often has complex characteristics, including seasonality, trends, and noise. It can be challenging for an autoencoder to capture and represent these complex features, especially for large-scale and high-dimensional time series data.

\subsection{Diffusion-based generation}
\label{app:diff}
The advantages of diffusion-based models are as follows. (i) Diffusion models make many record-making performances in different kinds of applications showing the power of such new deep generative models. (ii) The workflow of diffusion models is concise and flexible, making it possible to combine diffusion models with other generative models to improve the generative power, such as PDAE \cite{zhang2022unsupervised}. (iii) Diffusion models are usually straightforward to define and efficient to train. However, there are also disadvantages and challenges to overcome. Although the original definition of diffusion model is concise, it is not efficient when doing sampling and making likelihood estimations. There is still much work to do to make diffusion models more powerful.

\subsection{Advantages and disadvantages of three generative-based submethods}
\label{app:gensummary}
The details are shown in Table. \ref{table-gen-summary}.

\section{Advantages and Disadvantages of RNN, CNN, and GNN}
\label{app:rnncnngnn}

The main advantages of RNN, CNN, and GNN are as follows:

\begin{itemize}[leftmargin=*]
    \item RNN: \textbf{(i) Long-term dependencies.} RNN is suitable for capturing long-term dependencies in time series as it maintains internal states (hidden states) to memorize past information.  \textbf{(ii) Adaptable to varying lengths.} RNN can handle variable-length sequences, making it suitable for dynamic-length time series. \textbf{(iii) Global context information extraction.} Considers the entire history at each time step, aiding in learning global information.

    \item CNN: \textbf{(i) Local pattern extraction.} CNN effectively captures local patterns and features in input series, especially with 1D-CNN for time series. (ii) \textbf{Computational efficiency.} Comparison with RNN, CNN exhibits higher computational efficiency when dealing with longer time series data.

    \item GNN: \textbf{(i) Adaptability to graph structures.} GNN is suitable for handling multivariate time series data, where nodes represent variables and edges represent relationships between variables. \textbf{(ii) Dynamic relationship capture.} GNN can capture dynamic relationships between nodes, suitable for complex interactions between nodes at different variables.

\end{itemize}

The disadvantages of RNN, CNN, and GNN are as follows:

\begin{itemize}[leftmargin=*]
    \item RNN: \textbf{(i) Vanishing/exploding gradients.} Faces challenges of vanishing or exploding gradients during training, especially with deeper networks and long series. \textbf{(ii) Computational efficiency.} Compared to CNN, RNN exhibits lower computational efficiency when dealing with large-scale datasets.

    \item CNN: \textbf{(i) Long-term dependencies.} CNN may struggle to capture long-term dependencies in longer time series due to limited convolutional kernel size. \textbf{(ii) Information loss.} With increasing layers, information may gradually be lost during the propagation process.

    \item GNN: \textbf{(i) Computational and storage complexity.} GNN may face computational and storage complexity issues when dealing with large-scale graphs, especially those with a large number of nodes and edges. \textbf{Fixed-size graphs.} For fixed-size graph structures, GNN may struggle to adapt to time series of varying lengths. \textbf{Hyperparameter sensitivity.} GNN's performance may be sensitive to hyperparameter selection and the specific structure of the graph.

\end{itemize}

\section{Advantages and Disadvantages of Five Contrastive-based Methods}

\subsection{Sampling contrast}
\label{app:sc}
The advantages of this category of methods are as follows. Sampling contrast follows the most commonly used assumption in time series analysis. It has a simple principle and can model local correlations well. However, its disadvantage lies in the possibility of introducing fake negative sample pairs when analyzing long-term dependencies, leading to suboptimal representations.

\subsection{Prediction contrast}
\label{app:predc}
Prediction contrast is a self-supervised task based on context prediction. By predicting future information, it can learn meaningful and informative representations of time series data, capturing important features and patterns within the data. It focuses more on the slow-changing trends in time series data and can extract slow features. However, prediction contrast primarily focuses on local information and may not accurately model long-term dependencies in time series data. It is also sensitive to noise and outliers, which can affect the model's representation capacity and generalization performance.

\subsection{Augmentation contrast}
\label{app:ac}
Augmentation contrast is an efficient and straightforward self-supervised task. Its greatest advantage is its ease of implementation and understanding, making it applicable to various types of time series modeling tasks. However, their disadvantages are as follows. (i) Handling temporal dependencies is also a challenge since the essence of augmentation contrast lies in differentiating the feature representations of positive and negative sample pairs, rather than explicitly capturing time dependencies. (ii) Selecting appropriate augmentation methods for time series data is a challenging problem. (iii) Sampling bias is another issue, as it can lead to the generation of false negative samples.

\begin{table*}[!t]
    \caption{Summary of characteristics and limitations of three SSL methods.}
    \label{table-all-summary}
    \centering
    \begin{tabular}{c|cc} \toprule\toprule
    Method & Characteristics & Limitations\\ 
    \midrule
    \multirow{2}{*}{Generative-based} 
    & Learn the distribution of the original data  & Sensitivity to rare instances\\
    & Obtain compressed representation of data & Lack of high-level semantic information \\ 
    \midrule
    
    \multirow{2}{*}{Contrastive-based}
    & Good performance in classification tasks & Sampling bias of negative samples \\
    & Light-weight model (Encoder-only) & Uncertainty in data augmentation \\

    \midrule
    \multirow{2}{*}{Adversarial-based}
    & Useful auxiliary strategies for obtaining robust representations & Unstable training and easy to collapse \\
    & Strong sample generation capability & Rarely used for feature extraction \\
    \bottomrule \bottomrule
\end{tabular}
\vspace{-10pt}
\end{table*}

\subsection{Prototype contrast}
\label{app:protc}
Prototype contrast introduces the concept of prototypes, allowing samples to be assigned to a finite number of classes. Moreover, prototype contrast leverages high-level semantic information to encourage samples to exhibit clustered distributions in the feature space rather than uniform distributions, which is more consistent with real data distribution. However, the main issue with prototype contrast is that the number of prototypes needs to be predetermined, which still requires some prior information.

\subsection{Expert knowledge contrast}
\label{app:ekc}
This category is a relatively new self-supervised contrastive strategy. Its main feature is the incorporation of prior knowledge during the training process to guide the selection of positive and negative samples or the measurement of similarity. Its major advantage lies in the ability to accurately select positive and negative samples. However, its limitation lies in the requirement of providing reliable prior knowledge. In most scenarios, obtaining reliable prior knowledge for time series data is not easy. Incorrect or misleading knowledge can lead to biased representations.

\section{Advantages and Disadvantages of Two Adversarial-based Methods}

\subsection{Time series generation and imputation}
\label{app:tsgi}
The advantages of adversarial-based methods are as follows. GANs can generate high-quality time series samples. By learning the distribution of the data, GANs can perform imputation or generation task according to the seasonality and trends of different time series data, thereby improving the coherence and rationality of the results. Furthermore, there have been many efficient adversarial-based methods proposed in the field of image generation. It is possible that these methods can be transferred and applied to time series data generation or imputation tasks. However, the training process of GANs is relatively complex, requiring a trade-off between the generator and the discriminator, as well as addressing issues such as mode collapse. Since the generator is learned through training, if the training dataset is insufficient or of poor quality, the imputation or generation results may lack consistency in some specific cases. In extreme situations, the generated results may exhibit abnormalities or appear unreasonable.

\subsection{Auxiliary representation enhancement}
\label{app:are}
Adversarial strategies can help the model learn more robust representations, thereby improving the model's generalization ability. By introducing adversarial signals, the model can better adapt to the training data and resist interference or attacks. However, introducing adversarial strategies as a regularization term in the loss function increases the complexity of the training process. The competition between training the generator and the discriminator requires careful balancing, which may require more training time and computational resources. This can even lead to training instability. 

\section{Characteristics and limitations of three SSL methods}
\label{app:charlimi}
The details are shown in Table~\ref{table-all-summary}.

\section{Evaluation metrics, Examples, and Task Flow}
\label{app:app}

\subsection{Anomaly detection}
\label{app:ad}

\begin{figure}[!t]
	\centering
	\subfloat[Anomaly detection]{\includegraphics[width=80mm]{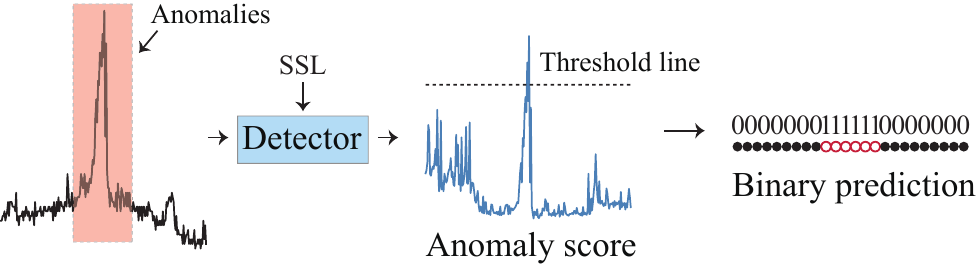}} \\
	\subfloat[Forecasting]{\includegraphics[width=80mm]{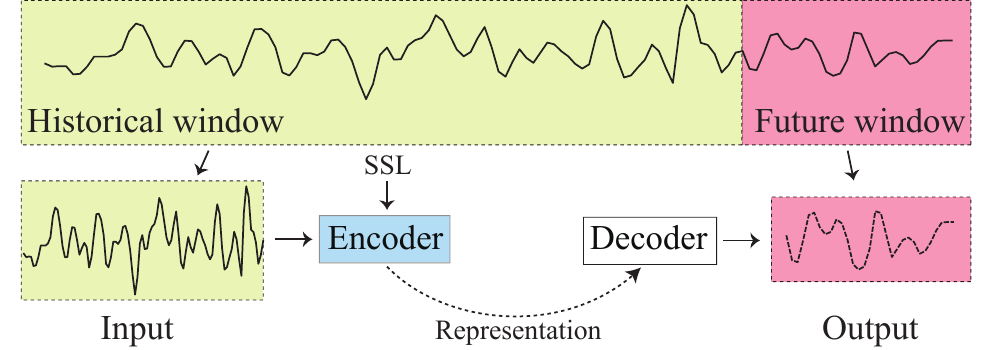}} \\
	\subfloat[Classification and clustering]{\includegraphics[width=80mm]{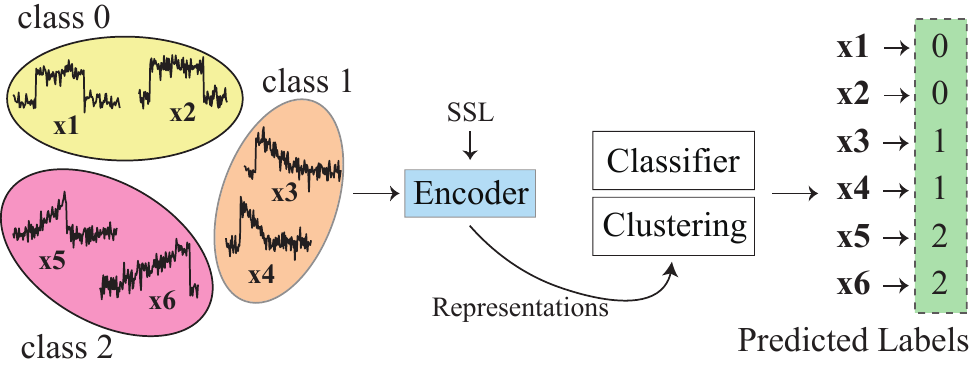}} \\
	\caption{Examples and task flows. The blue boxes represent modules where SSL can be deployed.}
	\label{fig:taskflow}
\vspace{-10pt}
\end{figure}

\begin{itemize}[leftmargin=*]

\item \textbf{Evaluation metrics.} Anomaly detection is essentially a binary classification problem, that is, determining whether a time step is normal or anomaly, so the most commonly used evaluation metrics include precision ($P$), recall ($R$), and F1-score ($F1$) \cite{sb2023navigating}. Note that in time series anomaly detection task, the adjustment strategy is widely adopted: if a time point in a certain successive abnormal segment is detected, all anomalies in this abnormal segment are viewed to be correctly detected \cite{xu2022anomaly,10.1145/3292500.3330672,DCdetector2023}.

Precision is the fraction of anomalous predictions that are actual anomalies, i.e.,
\begin{equation}
\label{p}
    P = TP/\left( {TP + FP} \right),
\end{equation}
Recall is the fraction of true anomalies that are correctly classified, i.e.,
\begin{equation}
\label{r}
    R = TP/\left( {TP + FN} \right),
\end{equation}
F1-score is the harmonic mean of precision and recall, i.e.,
\begin{equation}
\label{f1}
    F1 = (2 \cdot P \cdot R)/\left( {P + R} \right),
\end{equation}
where $TP$ is True Positive which represents the number of points that are labeled and predicted as anomalies. $FP$ is False Positive which represents the number of points that the predicted labels are anomalous, but the actual labels are normal. Contrary to $FP$, $FN$ is False Negative which denotes the number of points that the predicted labels are normal, but the actual labels are anomalous. Generally, a high $P$ means that the predicted anomalous points are correct and a high $R$ means that most real anomalous points are found, so both $P$ and $R$ should be high. In addition to the above metrics, there are also many available metrics in time series anomaly detection tasks, such as point-wise f-score, temporal distance, area under the curve, and etc. More details can be found in \cite{sb2023navigating}.

\item \textbf{Example and task flow}. Fig. \ref{fig:taskflow}(a)gives a time series example containing anomalies and the basic task flow for detecting anomalies. As shown in Fig. \ref{fig:taskflow}(a), the detector first converts input time series sample into the anomaly score, and then generates binary predictions by comparing the anomaly score to the given threshold. Finally, anomalies can be annotated through the comparison results.

\end{itemize}

\subsection{Forecasting}
\label{app:f}
\begin{itemize}[leftmargin=*]

\item \textbf{Evaluation metrics.} For the time series forecasting task, the (MSE) and mean absolute error (MAE) are widely used, i.e.,
\begin{equation}
    MSE = \frac{1}{m} \sum_{i=1}^{m}(x_i - \hat{x}_i)^2,
\end{equation}
\begin{equation}
\label{f1}
    MAE = \frac{1}{m} \sum_{i=1}^{m} \left| x_i - \hat{x}_i \right|.
\end{equation}
where $x_i$ is the true value and $\hat{x}_i$ is the predicted value.

\item \textbf{Example and task flow}. Fig. \ref{fig:taskflow}(b) gives a time series example and the task flow of time series forecasting. As shown in Fig. \ref{fig:taskflow}(b), a time series sample is first divided into the historical window and the future window, and then the context information of the historical window is encoded into the feature space through an encoder, and then the representation containing the historical information is decoded through the decoder to output the forecasting value of the future window. Here, SSL can help encoders obtain better historical window representations. 

\end{itemize}

\subsection{Classification and clustering}
\label{app:cc}

\begin{itemize}[leftmargin=*]

\item \textbf{Evaluation metrics.} For the time series classification and clustering task, the accuracy (Acc) is the widely used metric. Note that there is a difference in the accuracy calculation for classification and clustering. The clustering accuracy is defined as
\begin{equation}
Acc_{u} = \mathop {\max }\limits_{perm \in \mathcal{P}} \frac{1}{n}\sum\limits_{i = 0}^{n - 1} {\mathbbm{1}_{[ {perm\left( {{{\hat y}_i}} \right) = {y_i}} ]}}, 
\end{equation}
where $P$ is the set of all permutations in $\left[ {1,2, \ldots ,K} \right]$ where $K$ is the number of clusters. Alternatively, the classification accuracy is defined as
\begin{equation}
Ac{c_{s}}{\rm{ = }}\frac{{\rm{1}}}{n}\sum\limits_{i = 0}^{n - 1} {\mathbbm{1}_{[\hat{y}_i = y_i]}},
\end{equation}
where $\mathbbm{1} \in \left\{0,1\right\}$ is the indicator function evaluating to 1 if $\hat{y}_i = y_i$. $\hat{y}_i$ is predicted label and $y_i$ is the true label of $i$-th time series sample.

\item \textbf{Example and task flow}. Fig. \ref{fig:taskflow}(c) provides six time series samples and illustrates the task flow of classification and clustering. These six samples belong to three different classes. In the classification and clustering tasks, the classifier and clustering algorithm take these six samples as inputs and finally output the predicted label of each sample. SSL is usually applied in the training phase of the encoder to make the extracted representations more suitable for downstream label prediction tasks.

\end{itemize}

\begin{table*}[h]
    \caption{Quantitative comparisons of time series anomaly detection task.}
    \label{tab:adcompare}
    \centering
    \setlength{\tabcolsep}{1.8mm}{
    \begin{tabular}{c|ccc|ccc|ccc|ccc|ccc} \toprule\toprule
    \multirow{2}{*}{Method} & \multicolumn{3}{c|}{PSM} & \multicolumn{3}{c|}{SMD} & \multicolumn{3}{c|}{MSL} & \multicolumn{3}{c|}{SMAP} & \multicolumn{3}{c}{SWaT} \\ \cmidrule(lr){2-16}
     & P & R & F1 & P & R & F1 & P & R & F1 & P & R & F1 &P & R & F1 \\ \midrule
    THOC \cite{NEURIPS2020_97e401a0} &88.14  &90.99&89.54&79.76&90.95&84.99&88.45&90.97&89.69&92.06&89.34&90.68& 98.08 & 79.94 & 88.09        \\ 
    RANSynCoder \cite{10.1145/3447548.3467174} &99&	96&	93&	90&	85&	83& - &-&-&-&-&-& 95&	75&	84\\
    USAD \cite{10.1145/3394486.3403392} &-&-&-&93.14	&96.17	&93.82&	88.1&	97.86&	91.09&	76.97&	98.31	&81.86&	98.7 &74.02&	84.6\\
    Interfusion\cite{10.1145/3447548.3467075} &83.61&83.45&83.52&87.02&85.43&86.22&81.28&92.70&86.62&89.77&88.52&89.14&80.59&85.58&83.01\\
    OmniAnomaly \cite{10.1145/3292500.3330672} &88.39&74.46&80.83& 83.34&	94.49&	88.57&	88.67&	91.17&	89.89&	74.16&	97.76&	84.34 &81.42 & 84.30 & 82.83\\
    GRELEN \cite{ijcai2022p332}&94.2&92.1	&93.1&	88.2&	95.1&	91.5&-&-&-&-&-&-&			95.6	&83.5	&89.1\\
    VGCRN \cite{pmlr-v162-chen22x} &-&-&-&95&88.3&91.5&88.8	&94.1&91.4	&91.6	&92	&91.4&-&-&-\\
    ImDiffusion\cite{chen2023imdiffusion}& 98.11&97.53&97.81&95.2&95.09&94.88&89.3&86.38&87.79&87.71&96.18&91.75&89.88&84.65&87.09 \\
    Skip-CPC\cite{kexinnipsws2022} &98.36	&98.74&	98.55&	91.75&	97.34&	94.46	&90.84&	94.73&	92.75 &-&-&-&-&-&-\\
    TS-CP$^2$\cite{10.1145/3442381.3449903} &82.67&78.16&80.35&87.42&66.25&75.38&86.45&68.48&76.42&87.65&83.18&85.36 &-&-&- \\
    Dcdetector\cite{DCdetector2023} & 97.14&	98.74&	97.94&	83.59&	91.1&	87.18&	93.69&	99.69&	96.6&	95.63&	98.92&	97.02&	93.11&	99.77&	96.33\\
    AnomalyTrans \cite{xu2022anomaly}& 96.91& 98.9&	97.89&	89.4&	95.45&	92.33&	92.09&	95.15&	93.59&	94.13&	99.4&	96.69&	91.55&	96.73&	94.07 \\
    DAEMON \cite{10.1145/3539597.3570371} &-&-&-& 96.3&	96.2&	96.3&	91&	1	&95.3&	92.9&	89.2&	91&	96.6&	92.9&	94.7\\
    BeatGAN \cite{ijcai2019p616} &90.30&93.84&92.04&72.90&84.09&78.10&89.75&85.42&87.53&92.38&55.85&69.61&64.01&87.46&73.92 \\
    
    \bottomrule \bottomrule
\end{tabular}

}
\vspace{-10pt}
\end{table*}

\begin{table}[!t]
    \caption{Quantitative comparisons of time series forecasting task.}
    \label{tab:fcompare}
    \centering
    \setlength{\tabcolsep}{0.7mm}{
    
    \begin{tabular}{c|c|cc|cc|cc|cc} \toprule\toprule
    \multirow{2}{*}{Method} & \multirow{2}{*}{H} & \multicolumn{2}{c|}{ETTh} & \multicolumn{2}{c|}{ETTm} & \multicolumn{2}{c|}{Electricity} & \multicolumn{2}{c}{Weather}  \\ \cmidrule(lr){3-10}
    & & MSE & MAE & MSE & MAE & MSE & MAE & MSE & MAE\\ \midrule
    
    \multirow{2}{*}{LaST\cite{wang2022learning}} 
    & 24 &0.324 &0.368 &0.218 &0.289 &0.125 &0.222 &0.105 &0.134 \\
    & 48 &0.351 &0.380 &0.280 &0.329 &0.146 &0.245 &0.131 &0.174 \\ \midrule
    
    \multirow{2}{*}{CoST\cite{woo2022cost}}  
    & 24 &0.386 &0.429 &0.246 &0.329 &0.136 &0.242 &0.298 &0.360\\
    & 48 &0.437 &0.464 &0.331 &0.386 &0.153 &0.258 &0.359 &0.411\\ \midrule

    \multirow{2}{*}{USRL\cite{NEURIPS2019_53c6de78}}
    & 24 &0.942 &0.729 &0.689 &0.592 &0.564 &0.578 &0.522 &0.533\\
    & 48 &0.975 &0.746 &0.752 &0.624 &0.569 &0.581 &0.539 &0.543\\ \midrule

    \multirow{2}{*}{CPC\cite{DBLP:journals/corr/abs-1807-03748}}
    & 24 &0.728 &0.600 &0.478 &0.459 &0.403 &0.459 &0.328 &0.383\\
    & 48 &0.774 &0.629 &0.641 &0.550 &0.424 &0.473 &0.390 &0.433\\ \midrule

    \multirow{2}{*}{BTSF\cite{pmlr-v162-yang22e}}  
    & 24 & 0.541 &0.519 &0.302 &0.342 &-&- &0.324 &0.369\\
    & 48 & 0.613 &0.524 &0.395 &0.387 &-&- &0.366 &0.427\\ \midrule

    \multirow{2}{*}{TS-TCC\cite{ijcai2021-324}}  
    & 24 &0.653 &0.610 &0.473 &0.490 &0.311 &0.396 &0.372 &0.404\\
    & 48 &0.720 &0.693 &0.671 &0.665 &0.326 &0.407 &0.418 &0.445\\ \midrule

    \multirow{2}{*}{TS2Vec\cite{Yue_Wang_Duan_Yang_Huang_Tong_Xu_2022}} 
    & 24 &0.590 &0.531 &0.453 &0.444 &0.287 &0.375 &0.170 &0.309\\
    & 48 &0.624 &0.555 &0.592 &0.521 &0.309 &0.391 &0.231 &0.375\\ \midrule

    \multirow{2}{*}{TNC\cite{DBLP:journals/corr/abs-2106-00750}} 
    & 24 &0.708 &0.592 &0.522 &0.472 &0.354 &0.423 &0.200 &0.312\\
    & 48 &0.749 &0.619 &0.695 &0.567 &0.376 &0.438 &0.284 &0.367\\ \midrule

    \multirow{2}{*}{D$^3$VAE\cite{li2023generative}} 
    & 8  &0.292 &- &0.527 &- &0.251 &- &0.169 &-\\
    & 16 &0.374 &- &0.968 &- &0.308 &- &0.187 &-\\ \midrule

    \multirow{2}{*}{TimeGrad\cite{rasul2021autoregressive}} 
    & 8  &4.259 &- &1.877 &- &2.703 &- &2.715 &-\\
    & 16 &1.332 &- &2.032 &- &2.770 &- &1.110 &-\\ 
    
    \bottomrule \bottomrule
    \end{tabular}
    }
    \vspace{-5pt}
\end{table}

\begin{table*}[h]
    \caption{Quantitative comparisons of time series classification and clustering task.}
    \label{tab:cccompare}
    \centering
    \setlength{\tabcolsep}{0.8mm}{
    
    \begin{tabular}{c|cccccccccccccccc} \toprule\toprule

     \multirow{2}{*}{Method} & \multicolumn{16}{c}{15 UCR Datasets} \\ \cmidrule(lr){2-17}
     
    &Beef 
    &ChCo
    &Coffee
    &CricketX 
    &DiPOA
    &DiPOC
    &ECG200
    &ECGF
    &Car
    &Ham
    &Herring
    &Wine
    &Meat
    &MiPOA
    &Wafer
    &Yoga
    \\ \midrule

    TimeNet\cite{DBLP:journals/corr/MalhotraTVAS17} &- &73.1 &- &70.0 &77.0 &81.2 &- &92.6 &- &- &- &- &- &79.0 &- &84.0\\
    DCTR\cite{NEURIPS2019_1359aa93} &80.5 &53.6 &92.9 &- &78.2 &60.8 &66.5 &96.4 &75.1 &53.6 &57.6 &62.7 &97.6 &79.8 &73.4 & - \\
    USRL\cite{NEURIPS2019_53c6de78} &73.3 &78.2 &100.0 &77.7 &74.8 &77.5 &94.0 &100.0 &85.0 &72.4 &60.9 &87.0 &95.0 &66.2 &99.5 &87.8\\
    SSTSC\cite{XI2022105331} &- &- &- &68.9 &- &- &82.3 &- &- &- &60.2 &- &- &- &- &91.7\\
    TS2Vec\cite{Yue_Wang_Duan_Yang_Huang_Tong_Xu_2022} &76.7 &83.2 &100.0 &80.5 &77.0 &77.5 &94.0 &100.0 &88.3 &75.2 &64.1 &88.9 &96.7 &65.6 &99.8 &88.7\\
    DVSL \cite{10.1145/3340531.3412099} &90.0 &77.4 &100.0 &- &- &- &83.5 &97.4 &83.5 &- &65.6 &65.0 &98.8 &58.2 &- & -\\
    SSP-TSC \cite{9533426} &80.0 &79.2 &100.0 &82.1 &80.6 &80.4 &90.0 &98.5 &90.0 &80.0 &71.9 &94.4 &100.0 &66.2 &99.7 &89.1\\
    
    TNC \cite{DBLP:journals/corr/abs-2106-00750} &73.3 &76.0 &100.0 &62.3 &74.1 &75.4 &83.0 &83.0 &68.3 &75.2 &59.4 &75.9 &91.7 &64.3 &99.4 &81.2 \\
    MVTS-Trans\cite{10.1145/3447548.3467401} &50.0 &56.2 &82.1 &38.5 &74.1 &72.8 &99.9 &76.3 &55.0 &52.4 &59.4 &50.0 &90.0 &61.7 &99.1 &83.0 \\
    MuCL \cite{WICKSTROM202254} &67.0 &66.0 &94.0 &71.0 &62.0 &66.0 &87.0 &94.0 &78.0 &57.0 &57.0 &61.0 &84.0 &48.0 &99.0 & 79.0 \\
    \midrule

    \multirow{2}{*}{Method}& \multicolumn{16}{c}{HAR dataset and 14 UEA Datasets} \\ \cmidrule(lr){2-17}
    
    &HAR 
    &AWR
    &AF
    &CT
    &FD
    &HMD
    &Hb
    &IWB
    &JV
    &LSST
    &MI
    &NATOPS
    &P-S
    &PD
    &EW
    &FM
    
    \\ \midrule
    USRL\cite{NEURIPS2019_53c6de78} &- &98.7 &20.0 &99.4 &52.9 &35.1 &75.6 &16.0 &98.9 &55.8 &58.0 &94.4 &68.8 &98.5 &87.8 & 58.0 \\
    TS2Vec\cite{Yue_Wang_Duan_Yang_Huang_Tong_Xu_2022} &90.5 &98.7 &20.0 &99.5 &50.1 &33.8 &68.3 &46.6 &98.4 &53.7 &51.0 &92.8 &68.2 &98.9 &84.7 &48.0\\
    TNC \cite{DBLP:journals/corr/abs-2106-00750} &- &97.3 &13.3 &96.7 &53.6 &32.4 &74.6 &46.9 &97.8 &59.5 &50.0 &91.1 &67.6 &97.9 &84.0 &47.0\\
    TS-TCC\cite{ijcai2021-324} &89.2 &95.3 &26.7 &98.5 &54.4 &24.3 &75.1 &26.4 &93.0 &47.4 &61.0 &82.2 &73.4 &97.4 &77.9 &46.0\\
    ExpCLR\cite{pmlr-v162-nonnenmacher22a} &91.2 &- &- &- &- &- &- &- &- &- &- &- &- &- &- &-\\
    MVTS-Trans\cite{10.1145/3447548.3467401} &- &97.7 &6.7 &97.5 &53.4 &24.3 &74.6 &10.5 &97.8 &40.8 &50.0 &65.6 &74.0 &56.0 &74.8 &56.0\\
    TF-C \cite{nips-tfc} &78.2 &- &- &- &- &- &- &- &- &- &- &- &- &- &- &- \\
    CLOCS \cite{pmlr-v139-kiyasseh21a} &47.3 &- &- &- &- &- &- &- &- &- &- &- &- &- &- &- \\
    TapNet\cite{Zhang_Gao_Lin_Lu_2020} &- &98.7 &33.3 &99.7 &55.6 &37.8 &75.1 &20.8 &96.6 &56.8 &59.0 &93.9 &75.1 &98.0 &- &- \\
    ShapeNet\cite{Li_Choi_Xu_S_Bhowmick_Chun_Wong_2021} &- &98.7 &40.0 & 98.0 &60.2 &33.8 &75.6 &25.0 &98.4 &59.0 &61.0 &88.3 &75.1 &97.7 &- &- \\
    MHCCL \cite{meng2022mhccl} &91.6 &- &- &- &- &- &- &- &- &- &- &- &- &98.7 &79.1 &52.1\\
    MuCL \cite{WICKSTROM202254} &- &97.0 &17.0 &98.0 &50.0 &35.0 &68.0 &0.0 &87.0 &44.0 &54.0 &82.0 &71.0 &97.0 &71.0 &61.0\\
    \bottomrule \bottomrule
\end{tabular}
}

\vspace{-10pt}
\end{table*}

\section{Quantitative Comparisons and Correlation Analysis}
\label{app:qca}

\subsection{Quantitative performance comparisons}
\label{app:qpc}

The most direct way to compare the performance of different methods is to collect experimental results using the same dataset. In this subsection, we collect and compare the performances of these methods according to their downstream tasks. For time series anomaly detection, five popular datasets, PSM, SMD, MSL, SMAP, and SWaT, have been used by most of the methods. Similarly, for time series forecasting, four datasets, ETTh, ETTm, Electricity, and Weather, have often been referred to for evaluation where these methods have outperformed the best end-to-end forecasting models. For time series classification and clustering, we have chosen a total of 30 datasets from HAR dataset \cite{Anguita2013APD}, the UCR archive \cite{8894743}, and the UEA archive \cite{bagnall2018uea}, including Beef, ChlorineConcentration (ChCo), Coffee, CricketX, DistalPhalanxOutlineAgeGroup (DiPOA), DistalPhalanxOutlineCorrect (DiPOC), ECG200, ECGFiveDays (ECGF), Car, Ham, Herring, Wine, Meat, MiddlePhalanxOutlineAgeGroup(MiPOA), Wafer, Yoga, HAR, ArticularyWordRecognition (AWR), AtrialFibrillation (AF), CharacterTrajectories (CT), FaceDetection (FD), HandMovementDirection (HMD), Heartbeat (Hb), InsectWingbeat (IWB), JapaneseVowels (JV), LSST, MotorImagery (MI), NATOPS, PEMS-SF (PS), PenDigits (PD), EigenWorms (EW), and FingerMovements (FM). The details of these datasets can be found in Table \ref{tab:appdata}.

Table \ref{tab:adcompare} highlights the performance of several methods on five datasets in time series anomaly detection task. According to the results, we have the following
observations. (i) It can be seen that generative and contrastive-based methods are the preferred choices for performing anomaly detection tasks. Furthermore, the latest research shows that contrastive-based methods, such as Skip-CPC \cite{kexinnipsws2022} and Dcdetector \cite{DCdetector2023}, have achieved the best performance on four datasets. (ii) Two key factors can promote more accurate detection of anomalies: global/local association modeling and regularization constraints. Skip-CPC \cite{kexinnipsws2022}, Dcdetector\cite{DCdetector2023}, and AnomalyTrans \cite{xu2022anomaly} follow the same assumption, i.e., anomalies are difficult to have a strong association with the whole time series. This assumption has been proven to be reasonable in the task of anomaly detection. Adversarial-based regularization constraints can amplify the score difference between anomalies and normal points, thereby enhancing the accuracy of anomalies identification from another perspective, such as Minmax learning \cite{xu2022anomaly,10.1145/3394486.3403392} and adversarial learning \cite{10.1145/3539597.3570371}. 

Table \ref{tab:fcompare} provides experimental results for some methods in time series forecasting tasks on four datasets. It is important to note that time series forecasting is a highly popular task, and as a result, there are numerous methods related to this task \cite{zhou2022fedformer,Zhou_Zhang_Peng_Zhang_Li_Xiong_Zhang_2021,zhou2022film,10.1145/3534678.3539396,wang2022learning,10.5555/3495724.3497159,liu2022pyraformer}. Since our focus is on SSL, the methods in Table \ref{tab:fcompare} are based on self-supervised representation learning, distinguishing them from directly end-to-end methods. It can be seen that LaST \cite{wang2022learning} and CoST \cite{woo2022cost} show significant advantages compared to other methods. We believe the key lies in extracting disentangled representations of time series data, typically associated with latent seasonal-trend representations.

Table \ref{tab:cccompare} presents experimental results on 30 datasets for classification and clustering tasks. It can be observed that the majority of methods are constructed based on SSCL. This is because the learning objective of instance discrimination contrastive learning aligns with the goal of classification tasks, which is to recognize the classes of samples. According to the results, no single method can achieve the best performance on all datasets. This indicates that extracting representations from time series data is complex and challenging, and cannot be accomplished with a unified model. Therefore, in real-world applications, it is essential to choose an appropriate model based on the characteristics of the data. Additionally, most SSCL-based classification models emphasize that data augmentation also has a direct impact on the final classification accuracy \cite{Yue_Wang_Duan_Yang_Huang_Tong_Xu_2022,WICKSTROM202254,ijcai2021-324}. On the other hand, introducing expert features and adversarial losses also benefits time series classification and clustering tasks, enabling the model to learn more robust representations \cite{pmlr-v162-nonnenmacher22a,NEURIPS2019_1359aa93}.

\subsection{Correlation analysis between methods and tasks}
\label{app:corranalysis}
Different methods have different characteristics and limitations and therefore may be suitable for different tasks. Here we briefly summarize the correlation between the three methods (Generative-based, contrastive-based, and adversarial-based) and the four tasks. In summary, generative-based SSL is more suitable for anomaly detection and forecasting tasks, while contrastive-based SSL is more suitable for classification and clustering tasks. This is because the essence of forecasting and anomaly detection tasks is the process of data generation, which is consistent with the core of generative-based SSL methods. Most of the pretext tasks of contrastive-based SSL are discriminative tasks, which is consistent with the goals of classification and clustering tasks. Adversarial-based methods are also widely used for forecasting and anomaly detection tasks, since most adversarial-based methods include a generator. On the other hand, another main role of adversarial-based methods is as an auxiliary regularization term to ensure that the features extracted by the model are more robust and informative. In addition to applying a kind of methods independently, existing results and work also show that a hybrid approach (mixing multiple SSL strategies) is often a better choice.

\end{document}